\documentclass[11pt]{article}
\usepackage{hyperref}
\usepackage{url}
\usepackage[utf8]{inputenc} 
\usepackage[T1]{fontenc}    
\usepackage{hyperref}       
\usepackage{url}            
\usepackage{booktabs}       
\usepackage{amsfonts}       
\usepackage{nicefrac}       
\usepackage{microtype}      
\usepackage{xcolor}         

\usepackage{microtype}
\usepackage{graphicx}
\usepackage{subfigure}
\usepackage{tcolorbox}
\usepackage{booktabs} 
\usepackage{hyperref}
\usepackage{url}
\usepackage{algorithm}
\usepackage{colortbl}
\usepackage{svg}
\usepackage{graphicx}
\usepackage{bbm}
\usepackage{booktabs} 
\usepackage{adjustbox} 
\usepackage{multirow} 
\usepackage{dsfont}
\usepackage{wrapfig}
\usepackage{caption}

\usepackage{graphicx}
\usepackage{hyperref}
\usepackage{amsmath}
\usepackage{amssymb}
\usepackage{mathtools}
\usepackage{amsthm}
\usepackage{enumitem}

\usepackage[capitalize,noabbrev]{cleveref}

\theoremstyle{plain}

\theoremstyle{definition}

\usepackage[preprint]{acl}

\usepackage{times}
\usepackage{latexsym}

\usepackage[T1]{fontenc}

\usepackage[utf8]{inputenc}

\usepackage{microtype}

\usepackage{inconsolata}

\usepackage{graphicx}

\usepackage[textsize=tiny]{todonotes}

\usepackage{subcaption}

\usepackage{tcolorbox}
\usepackage{titlesec}
\usepackage{wrapfig}  
\tcbuselibrary{listingsutf8}
\usepackage{subcaption}
\usepackage{subfig}
\tcbset{
  systembox/.style={
    colback=gray!10,
    colframe=gray!70,
    fonttitle=\bfseries,
    title=System Prompt,
    boxrule=0.5mm,
    arc=4mm,
    left=4mm,
    right=4mm,
    top=2mm,
    bottom=2mm
  },
  userbox/.style={
    colback=gray!5,
    colframe=gray!70,
    fonttitle=\bfseries,
    title=User Prompt,
    boxrule=0.5mm,
    arc=4mm,
    left=4mm,
    right=4mm,
    top=2mm,
    bottom=2mm
  }
}

\title{How Memory Management Impacts LLM Agents: \\ An Empirical Study of Experience-Following Behavior}

  \author{
\bf Zidi Xiong$^{1}$\thanks{Equal contributions. Correspondence to: Zidi Xiong \href{mailto:zidixiong@g.harvard.edu}{\texttt{zidixiong@g.harvard.edu}} and Zhen Xiang\href{mailto:zhen.xiang.lance@gmail.com}{\texttt{zhen.xiang.lance@gmail.com}}},
Yuping Lin$^{3*}$,
Wenya Xie$^{4*}$,
Pengfei He$^{3}$, \\
\bf
Zirui Liu$^{4}$,
Jiliang Tang$^{3}$, 
Himabindu Lakkaraju$^{1}$, 
Zhen Xiang$^{2}$ \\
  $^1$Harvard University \quad
  $^2$University of Georgia \\
  $^3$Michigan State University\quad
  $^4$University of Minnesota-Twin Cities
 \\
}

\makeatletter
\providecommand{\sf@counterlist}{} 
\makeatother

\begin{document}
\maketitle

\begin{abstract}
Memory is a critical component in large language model (LLM)-based agents, enabling them to store and retrieve past executions to improve task performance over time.
In this paper, we conduct an empirical study on how memory management choices impact the LLM agents' behavior, especially their long-term performance. 
Specifically, we focus on two fundamental memory management operations that are widely used by many agent frameworks—memory \textit{addition} and \textit{deletion}—to systematically study their impact on the agent behavior.
Through our quantitative analysis, we find that LLM agents display an \textit{experience-following property}: high similarity between a task input and the input in a retrieved memory record often results in highly similar agent outputs.
Our analysis further reveals two significant challenges associated with this property: \textit{error propagation}, where inaccuracies in past experiences compound and degrade future performance, and \textit{misaligned experience replay}, where some seemingly correct executions can provide limited or even misleading value as experiences.
Through controlled experiments, we demonstrate the importance of regulating experience quality within the memory bank and show that future task evaluations can serve as free quality labels for stored memory.
Our findings offer insights into the behavioral dynamics of LLM agent memory systems and provide practical guidance for designing memory components that support robust, long-term agent performance. 
We also release our code to facilitate further study. \footnote{The dataset and experimental code are available in \href{https://github.com/yuplin2333/agent_memory_manage.git}{\text{https://github.com/yuplin2333/agent\_memory\_manage.git}}}

\end{abstract}

\vspace{-0.05in}
\section{Introduction}
\label{intro}
\begin{figure}[h!]
    \vspace{-0.075in}
    \centering
    \includegraphics[width=\linewidth]{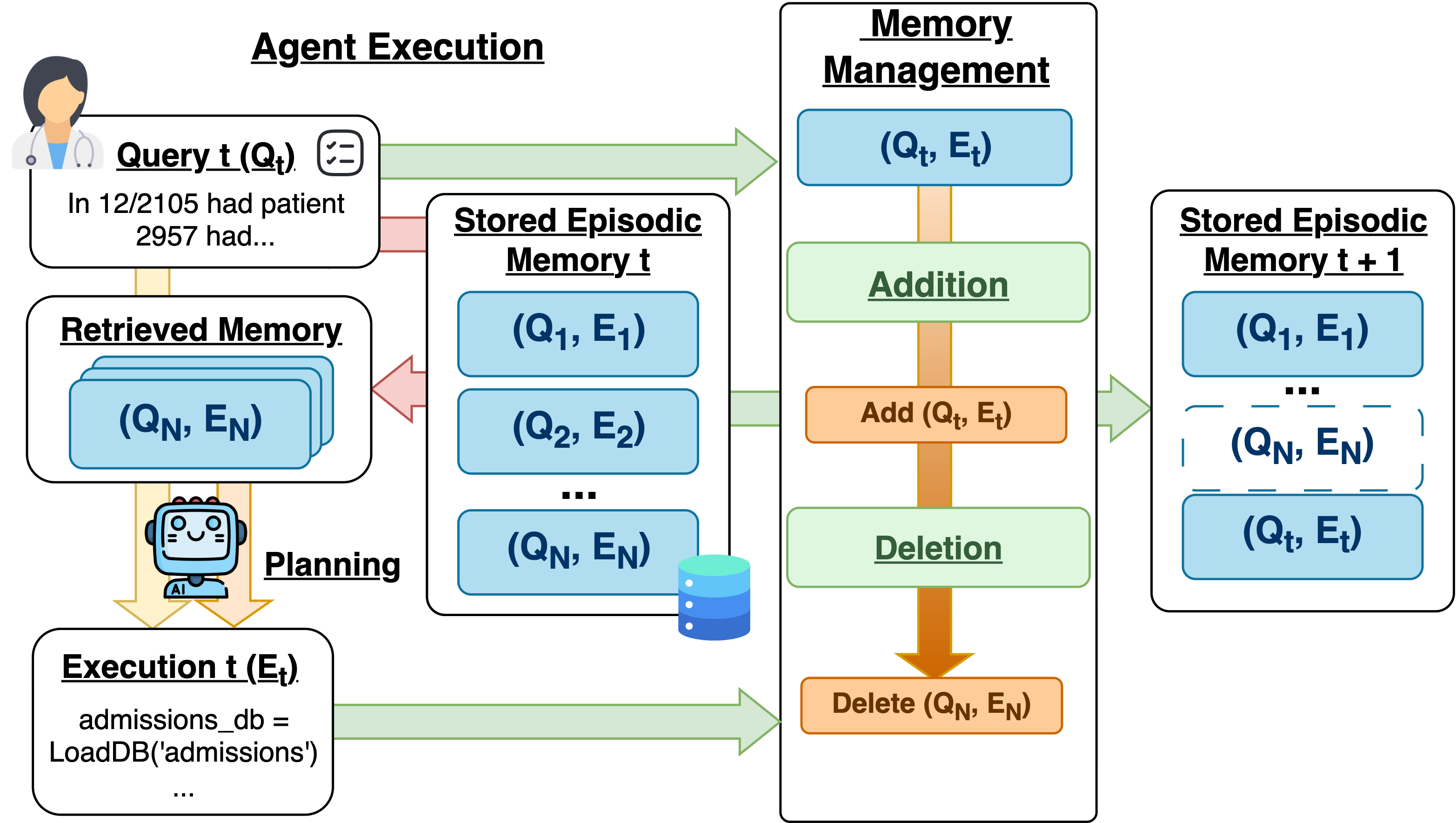}
    \vspace{-0.2in}
    \caption{
    Illustration of the memory management workflow after each agent execution.
    }
    \label{fig:teaser}
    \vspace{-0.3in}
\end{figure}


To enable effective solving of complex tasks and self-evolving over time, large language model (LLM) agents often equip an episodic memory module~\cite{wang2024survey}.
This dynamic mechanism retains past task queries (i.e. the agent input) and execution (i.e. the agent output), which can be retrieved as demonstrations to guide similar future tasks.
Through effective memory management, including adding, updating, and deleting past experiences, the performance of LLM agents can be improved over time~\cite{zhang2024survey}.

{
In practice, memory designs vary widely across agents due to their diverse objectives and functionalities~\cite{shi2024ehragent,mao2023language,wang2023voyager,li2023tradinggpt,park2023generative,xiang2024guardagent}.
To address this heterogeneity, recent studies have proposed a range of memory management strategies~\cite{yin2024explicit,zeng2024structural,zhao2024expel,zhong2024memorybank,xu2025mem}.
However, these approaches are often tailored to specific tasks and offer limited understanding of the underlying principles that govern memory behavior across different agentic systems.
}

{
In this work, we investigate a fundamental question in memory management: \textit{How do the evolving dynamics of the memory bank, driven by continuous memory operations, influence long-term agent execution?}
In particular, we focus on the core mechanisms underlying these dynamics—memory addition and memory deletion—as they constitute the foundation of memory management (see Figure~\ref{fig:teaser}).
Moreover, unlike prior studies on in-context learning with static, external knowledge bases~\citep{luo2024context}, our setting with these two operations captures the distinctive nature of agentic memory systems: a \textbf{dynamic} retrieval pool (memory bank) that evolves over time and contains inherently \textbf{noisy} outputs (trajectories) that are often generated by the agent itself. 
}
Through extensive experiments, we identify an important phenomenon that we term the \textit{experience-following} property: a high `input similarity' between the current task query and the one from the retrieved record often yields a high `output similarity' between their corresponding (output) executions.
{While this property enables effective reuse of successful experiences, we uncover two significant challenges arising from the dynamic and noisy nature of agentic memory banks.}
First, we observe the problem of \textit{error propagation}: if a retrieved memory record contains noisy or incorrect outputs, the agent is likely to replicate and even amplify these errors during the current task. If the resulting execution is then added back into memory, the error is likely to be further propagated to future tasks. 
{Second, we recognize the issue of \textit{misaligned experience replay},
which limits the benefits of experience following—certain memory records, when retrieved as demonstrations, consistently lead to poor execution due to their misalignment with the current task, indicating their inadequacy as effective demonstrations.
Retaining these records increases the likelihood of suboptimal or incorrect executions.}

{To systematically analyze these phenomena, we conduct controlled experiments across four distinct agents operating on diverse tasks. Crucially, we reveal the critical yet underexplored role of \textit{trajectory evaluators} in both memory addition and deletion processes, providing novel insights related to effective memory bank optimization that have been significantly overlooked in existing literature.}

We further examine challenging conditions representing real-world scenarios, including (1) \textit{Task Distribution Shift:} the task distribution changes substantially over time, requiring the agent to adapt to shifting patterns and contexts. (2) \textit{Memory Resources Constraint:} the memory capacity is severely limited, requiring the agent to retain only the most valuable and helpful experience. 
The results underscore the importance of thoughtful memory management for maintaining stable and effective long-term agent performance.

Our contributions are summarized below:

\begin{itemize}[leftmargin=*]
\vspace{-0.1in}
\setlength\itemsep{0em}
\item {We conduct systematic and quantitative analysis of the agent memory dynamics through two most essential memory operations—addition and deletion—of the memory records on the long-term performance of LLM agents.
Our findings provide key principles for robust memory management design.}
\item We reveal an experience-following property for LLM agents with memory and highlight two important challenges for memory management—error propagation and misaligned experience replay—arising with this property. 
\item {
We demonstrate how incorporating environmental feedback from trajectory evaluators can effectively improve memory management by mitigating both error propagation and misaligned experience replay.
}
\item We study how memory evolution affects long-term agent performance under task distribution shifts and memory constraints, finding that well-chosen evaluators enable agents to adapt to these scenarios and sustain performance with simple addition and deletion.
\end{itemize}

\section{Background and Related Works}
\label{related}

\subsection{Memory Module of LLM Agents}
LLM agents often include short-term memory and long-term memory~\cite{zhang2024survey}. Short-term memory usually refers to inside-task working memory~\cite{sumers2023cognitive}, while long-term memory~\cite{sumers2023cognitive} can be divided into three types: \textit{semantic memory}, \textit{procedural memory}, and \textit{episodic memory}. Semantic memory~\cite{Kumar2020SemanticMA} contains the agent's world knowledge and understanding of the environment; procedural memory~\cite{proceduralmemory} involves rules or procedures, which may reside implicitly in the LLM's weights or be explicitly defined as guidelines for the agent; and episodic memory~\cite{Nuxoll2007ExtendingCA,lampinen2025latentlearningepisodicmemory} records task-specific experiences. 
In this paper, we specifically focus on episodic memory, which is commonly used by LLM agents with memory and has recently been shown to mitigate the lack of latent learning capability in LLMs~\citep{lampinen2025latentlearningepisodicmemory}.

\subsection{Memory Management and Limitations}
LLM agents often employ an episodic memory module to store past experiences for future retrieval, which is crucial for effective planning and task execution~\cite{shi2024ehragent,mao2023language,xu2025mem,zhou2025agentfly,zhong2024memorybank,yin2024explicit}. 
The memory module typically includes operations such as memory reading and memory management~\cite{zhang2024survey}. 
Specifically, given a new task query $q$ and a memory base currently containing $N$ query-execution pairs $\mathcal{D} = \{(q_1, e_1), \ldots, (q_N, e_N)\}$, the agent's execution cycle involves the following memory operations:

\paragraph{Memory Reading:}  The agent retrieves a subset $\xi_K \subset \mathcal{D}$ consisting of the $K$ query-execution pairs most relevant to the query $q$. The relevance is often measured by the input similarity between the task query $q$ and the query from the retrieved past experience.
For example, the relevance can be computed as the cosine similarity between the feature representations for both queries from a text encoder~\cite{kusupati2022matryoshka}. The retrieved pairs $\xi_K$ are then used as in-context learning demonstrations to guide the LLM in generating an execution trajectory $e$ for the task query $q$.

\paragraph{Memory Management and Limitations:} 
Memory management typically involves the \textit{addition} and \textit{deletion} of memory records.
Specifically, when obtaining the query-execution trajectory pair $(q, e)$, addition decides whether this pair should be added to memory, while deletion~\cite{zhong2024memorybank,wang2024trove} is often triggered to remove outdated or redundant pairs from memory.

Although various strategies have been proposed for memory management, such as structural transformation~\cite{anokhin2024arigraph,zeng2024structural,xu2025mem}, merging~\cite{zhong2024memorybank,liu2023think,hu2024hiagent}, summarization~\cite{wang2024agent,zhong2024memorybank,pan2025memory}, and reflection~\cite{shinn2024reflexion,zhao2024expel,liang2024self}, 
these approaches are often designed for specific agent types (e.g., chatbot agents) and lack a unified design that generalizes across different agentic systems.

{Recent works also study the optimization of the memory bank, such as using an expectation–maximization approach for optimization~\cite{yin2024explicit} or formulating memory management as a Markov Decision Process~\citep{zhou2025agentfly}.
However, these works do not provide an understanding of how memory bank dynamics affect long-term performance under the inherently noisy nature of basic addition and deletion operations. In contrast, our findings highlight potential challenges posed by these operations and offer insights to guide the design of future memory systems.}

\section{Addition of Memory}
\label{sec:add}
Memory addition refers to the process of deciding whether a finished task execution should be stored as a new record in the memory bank. Here, we consider the most common approach that adds memory based on the feedback of the trajectory evaluator~\citep{wang2024agent,shi2024ehragent}. We investigate three types of memory addition with different trajectory evaluators—an add-all baseline, selective addition based on coarse automatic evaluation, and selective addition based on strict human evaluation—along with a fixed-memory baseline without memory addition. 
\subsection{Setup}
\label{sec:add_setup}

\paragraph{Setup for Agents} We investigate memory management in generic LLM agent settings by considering an illustrative and \textit{controllable} synthetic agent and three representative agents designed for different tasks:  
{For the synthetic agent, we design a task in which the agent predicts the output of a simple mapping -- a linear function in the ground truth -- by performing regression approximation over the provided demonstrations, which we call \textbf{RegAgent}. 
Formally, the agent receives an input vector $x$ together with several past guesses over nearby inputs under the linear transformation $w$ to infer an output for $w^{\top}x$.
This setup allows us to investigate an ideal setting where the model fully relies on the provided demonstrations to generate the output, while enabling control over the noise level of the stored memory and \textit{direct measurement} of the resulting error.}

For real agent, we include \textbf{EHRAgent}~\cite{shi2024ehragent}, \textbf{AgentDriver}~\cite{mao2023language}, and \textbf{CIC-IoT Agent}~\cite{ciciot}.
In addition to their distinct tasks, these agents also vary significantly in input-output formats and memory retrieval mechanisms, which enhances the generalizability of our findings.
For all four agents, we use GPT-4o-mini as the backbone language model for most of the experiments.
See Appendix~\ref{app:detail_functionality} for details of these agents.


\paragraph{Setup for Memory Addition} We begin by describing several memory addition using different trajectory evaluator.
For a query-execution pair \((q, e)\) and an evaluator \(\pi\), the addition decision is given by \(\pi(q, e)\): if \(\pi(q, e) = 1\), the experience is stored; if \(\pi(q, e) = 0\), it is discarded. Our aim is to analyze how the noise introduced by additions from different evaluators influences the agent's behavior and long-term performance.
Starting from an identical initial memory, we examine the following four categories of memory additions:

\textbf{1) Fixed-memory baseline:} 
In this baseline setting, we use a subset of the training data with correct agent execution as the fixed memory bank~\cite{mao2023language,li2023tradinggpt,zhao2024expel,zheng2023synapse}, and the agent relies only on this fixed memory without adding new entries, i.e., \(\pi_{\rm fixed}(q, e) = 0\).

\textbf{2) Add-all approach:} A straightforward addition is to store every encountered task and its execution: \(\pi_{\rm all}(q, e) = 1\).

\textbf{3) Selective addition based on automatic evaluation (coarse):}
{
We have \(\pi_{\rm automatic}(q, e) = \text{Automatic}(q, e)\) and evaluate three automatic evaluators~\citep{wang2024agent,yin2024explicit,liang2024self}, denoted as \underline{\textbf{Coarse 1 (C1)}}, \underline{\textbf{Coarse 2 (C2)}}, and \underline{\textbf{Coarse 3 (C3)}}.
For RegAgent, we define the evaluator based on the absolute error between the predicted and ground-truth outputs, where a smaller error threshold indicates a more accurate evaluator.
Specifically, we set the thresholds to 1.6, 1.4, and 1.2 for C1, C2, and C3, respectively.
For the other three agents, C1 and C2 correspond to GPT-4o-mini and GPT-4.1-mini used as evaluators, respectively, while C3 employs GPT-4.1-mini fine-tuned on 300 correct judge data collected from executions on a separate training set.
}
Detailed designs of these automatic evaluators are provided in Appendix~\ref{ap:coarse-evaluator-prompts}.

\textbf{4) Selective addition based on human evaluation (strict):}In this stricter approach, a human (oracle) serves as the evaluator~\citep{shi2024ehragent}, determining whether the execution should be stored in the memory: \(\pi_{\rm human}(q, e) = \text{Human}(q, e)\). Although human input could be gathered after each execution for practical agents, it is not feasible for our current evaluation. Therefore, we simulate this process by comparing the generated output with the ground truth. More details on each agent’s specific designs for memory addition can be found in Appendix~\ref{app:detailed_setup}.

\label{sec:addition}

\subsection{Execution quality and memory size jointly determine long-term agent performance.}
\label{sec:add_result}
\begin{table}[ht!]
\centering
\caption{Performance of the three memory addition strategies: add-all, coarse evaluation with three different automatic evaluators, and selective addition based on human evaluation (strict), compared with the fixed-memory baseline (shaded).
The results highlight the necessity of selection for effective memory addition.
}
\setlength{\tabcolsep}{2pt}
\begin{adjustbox}{width=\linewidth} %
\Large
\begin{tabular}{c|cc|cc|cc|cc}
\toprule
\multirow{2}{*}{\textbf{Judge}} & \multicolumn{2}{c|}{\textbf{RegAgent}}
& \multicolumn{2}{c|}{\textbf{EHRAgents}} 
& \multicolumn{2}{c|}{\textbf{AgentDriver}} 
& \multicolumn{2}{c}{\textbf{CIC-IoT Agent}} \\  
& \textbf{SR. $\uparrow$} & \textbf{Mem Size $\downarrow$}  
& \textbf{ACC $\uparrow$} & \textbf{Mem Size $\downarrow$}  
& \textbf{SR. $\uparrow$} & \textbf{Mem Size $\downarrow$} 
& \textbf{ACC. $\uparrow$}  & \textbf{Mem Size $\downarrow$} \\ \hline
\rowcolor{gray!20}\textbf{Fixed}  &67.53 & 100  &16.75              
 & {100} & 40.11 & {180} & 71.50 & 50   \\ \hline 
\textbf{Add all}     & 55.48 & 4100          
&13.05&  2411  & 32.32 & 2125  & 59.90 &  1050 \\ \hline
\rowcolor{gray!20}\multicolumn{9}{c}{\textbf{Coarse}}        \\     
\textbf{C1}  &63.18 &3511   &26.19&  1447& 36.92  &  1161 & 74.00 &  1030 \\ \hline
\textbf{C2}   &65.78 & 3347 & 32.21 & 1467 & 40.01 & \textbf{1119} & 68.80 &  936\\ \hline
\textbf{C3}  &67.35 & 3139  & 34.66 & 1094 & 47.37 & 1285 &79.50 & 952\\ \hline
\rowcolor{gray!20}\multicolumn{9}{c}{\textbf{Strict}}        \\
\textbf{Strict}     &\textbf{70.95} & \textbf{2938}               
&\textbf{38.50}& \textbf{1012}  & \textbf{51.00} &  {1178}   & \textbf{85.40} & \textbf{904} \\ \bottomrule
\end{tabular}
\end{adjustbox}
\label{tab:comparison}
\end{table}
\begin{figure}
    \centering
        \includegraphics[width=0.46\linewidth]{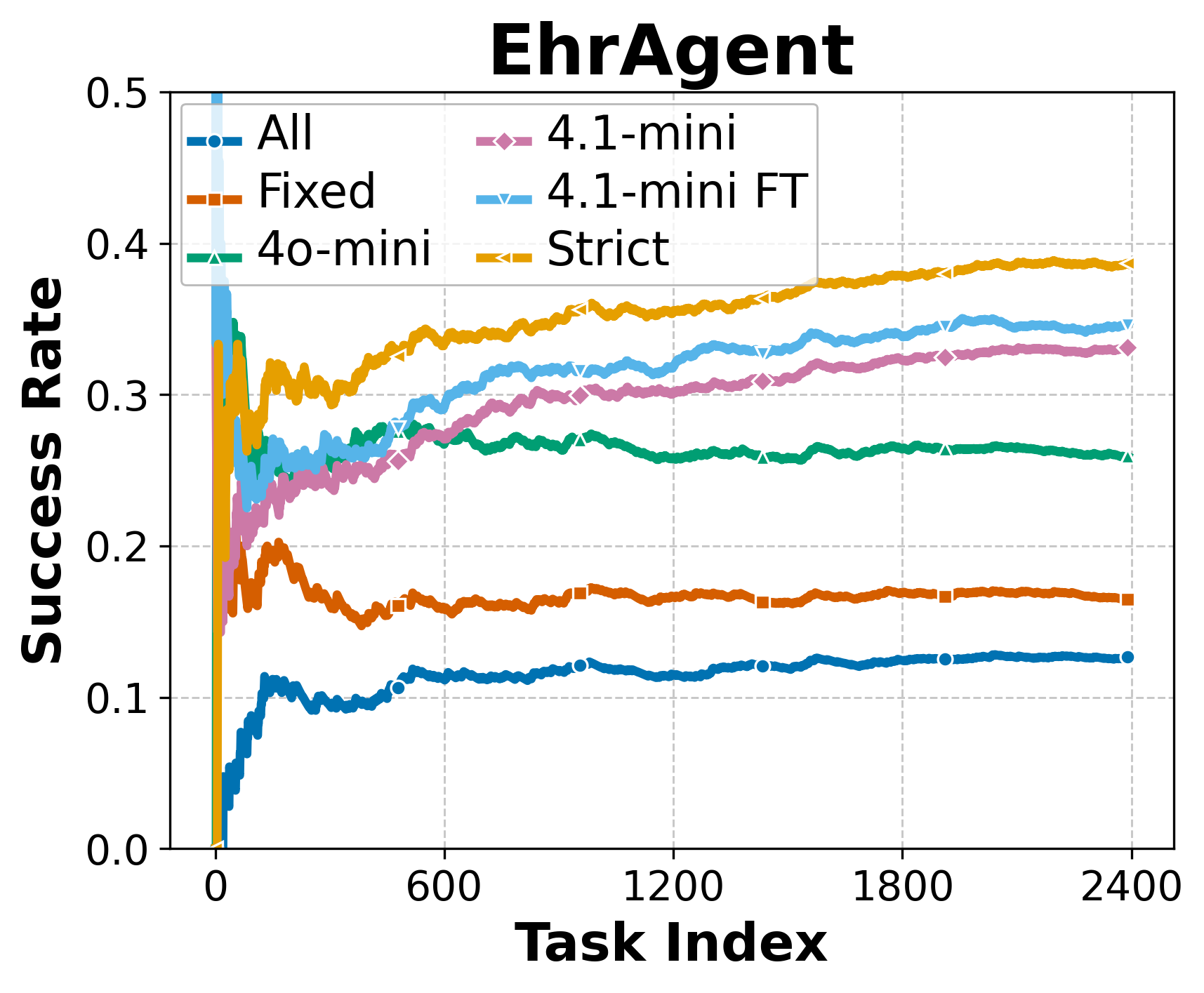}
          \includegraphics[width=0.46\linewidth]{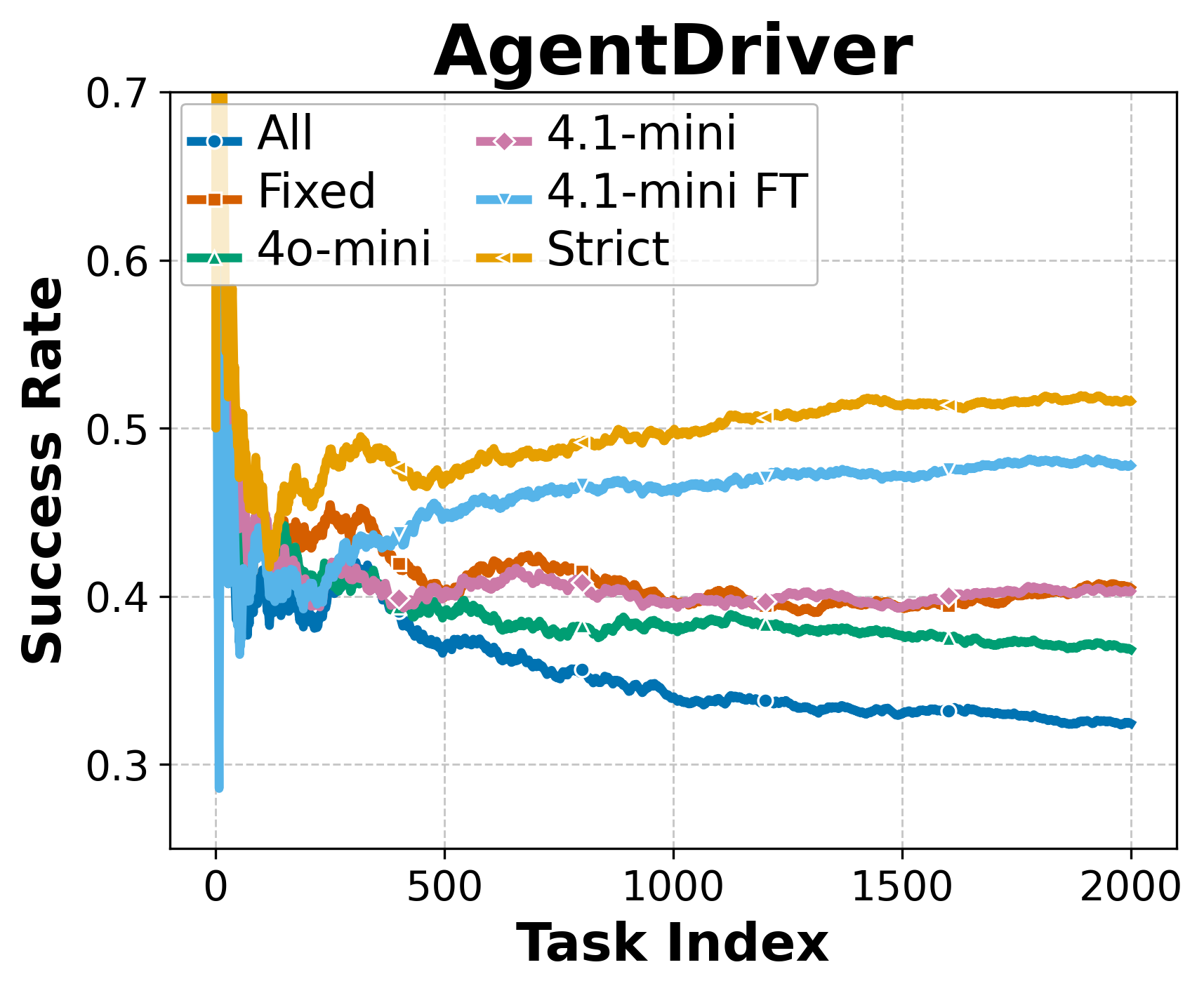} 
    \caption{Performance trend for EHRAgent and AgentDriver.
4o-mini, 4.1-mini, and 4.1-mini FT denote different coarse evaluators from the GPT series. Both the strict evaluator and some coarse evaluators exhibit consistent self-improvement over time.
    }
    \label{fig:ehr_add_acc}
     \vspace{-0.2in}
\end{figure}


\begin{figure*}[t!]
    \centering

    \begin{minipage}[t]{0.48\textwidth}
        \centering
        \includegraphics[width=0.48\linewidth]{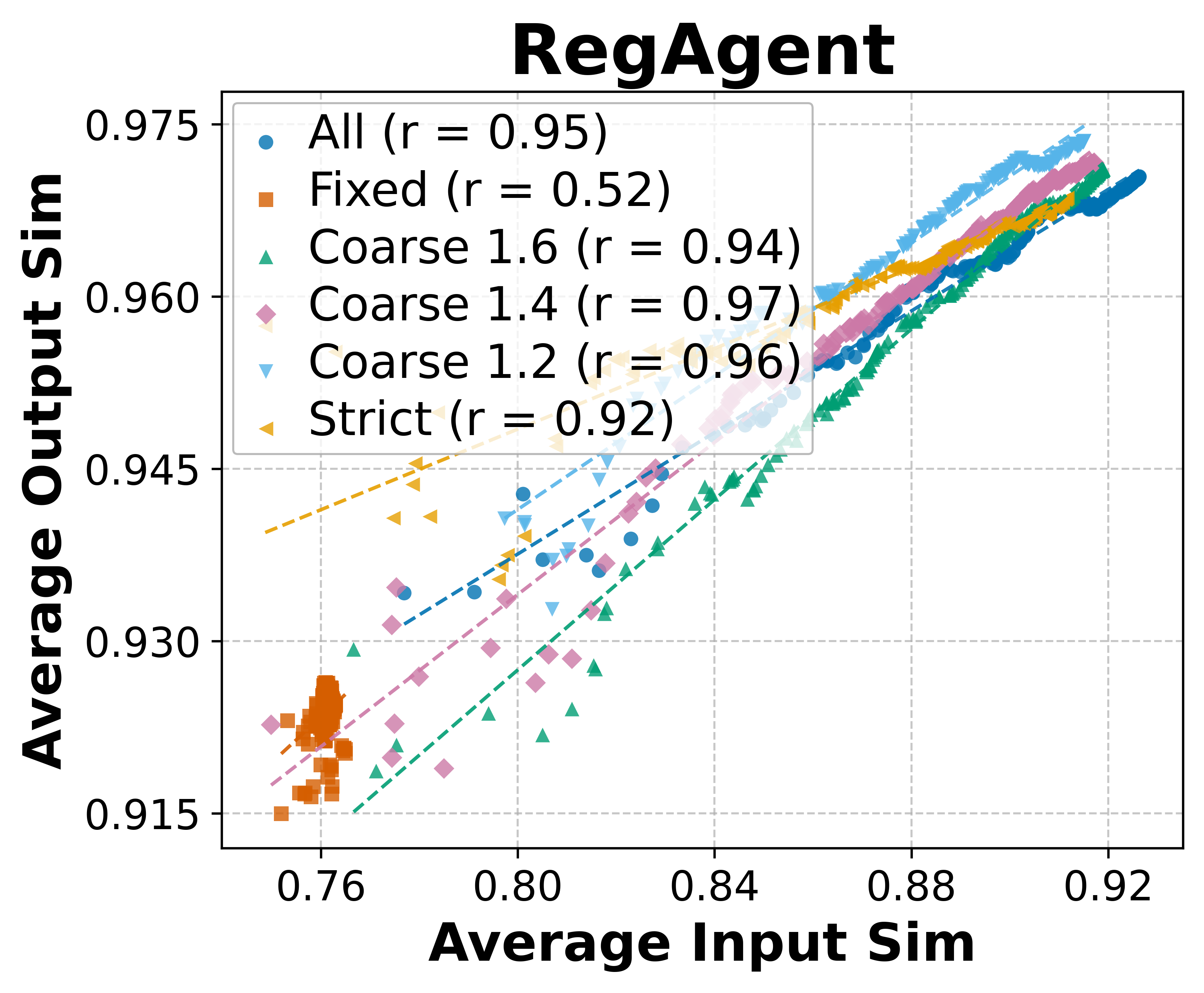}
        \includegraphics[width=0.48\linewidth]{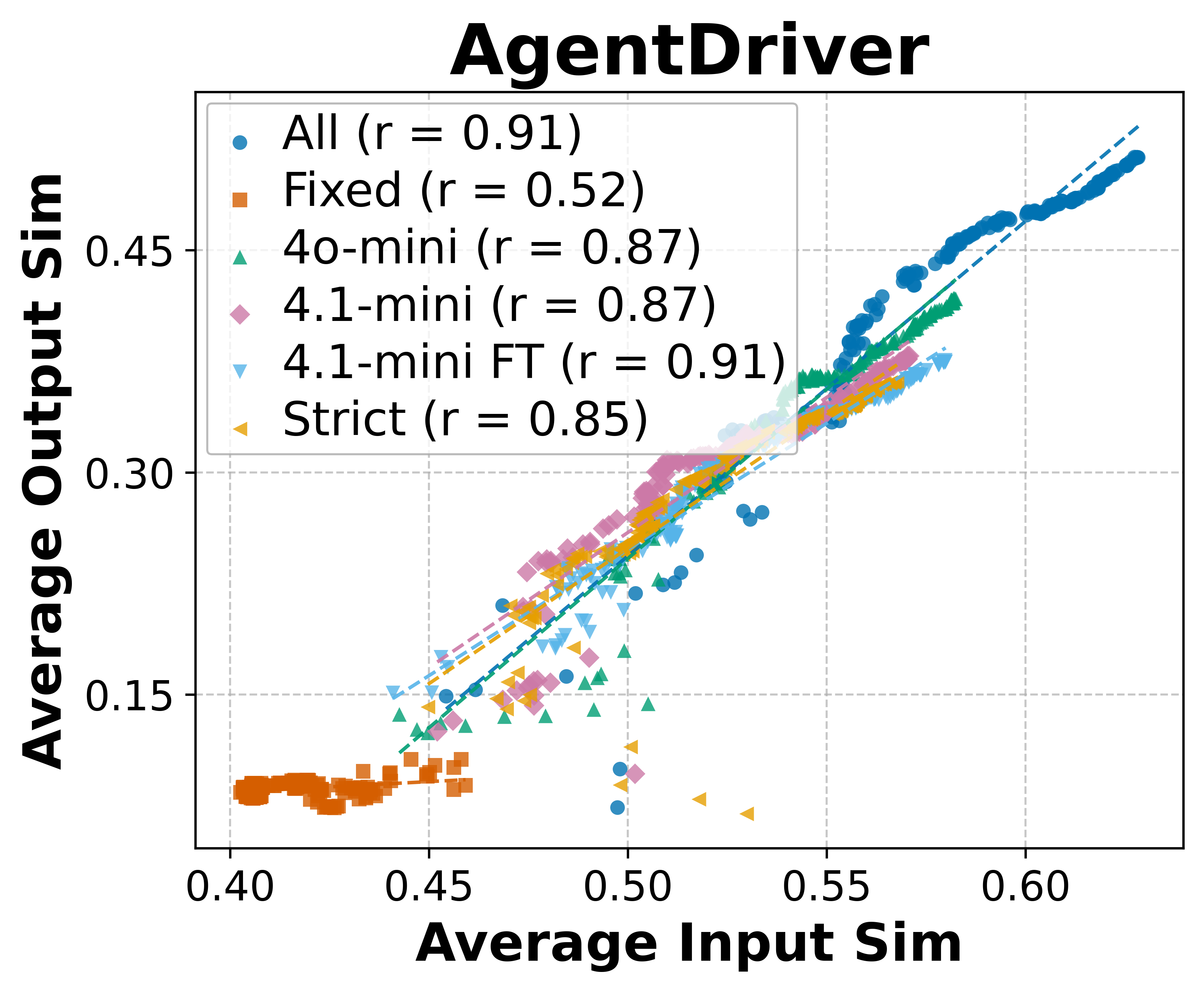}
        \caption{
        \textbf{Left:} Output similarity versus input similarity for RegAgent over different evaluators. 
        \textbf{Right:} Output similarity versus input similarity for AgentDriver over different evaluators.
        }
        \label{fig:ehr_add_sim}
    \end{minipage}
    \hfill
    \begin{minipage}[t]{0.48\textwidth}
        \centering
        \includegraphics[width=0.48\linewidth]{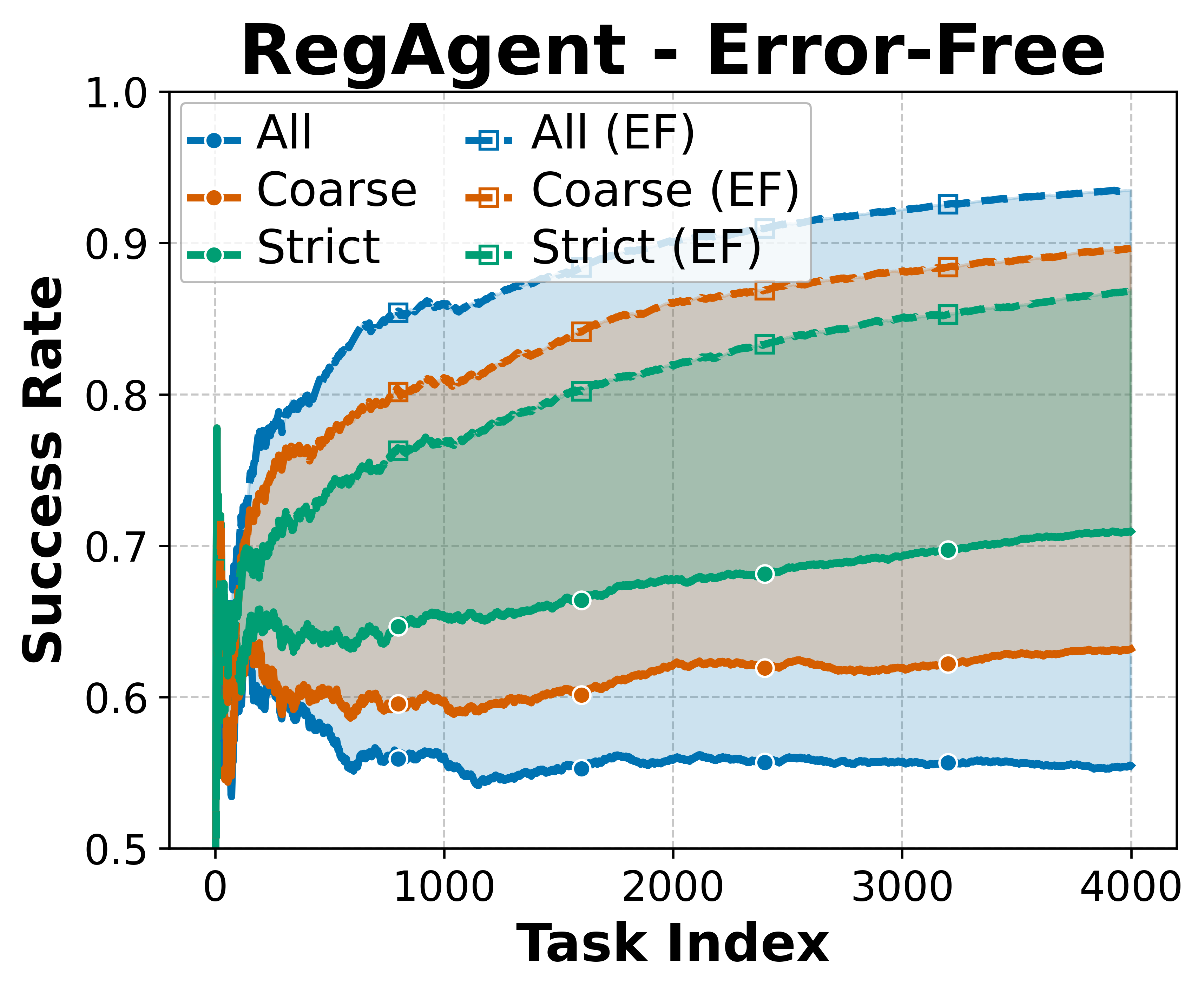}
                \includegraphics[width=0.48\linewidth]{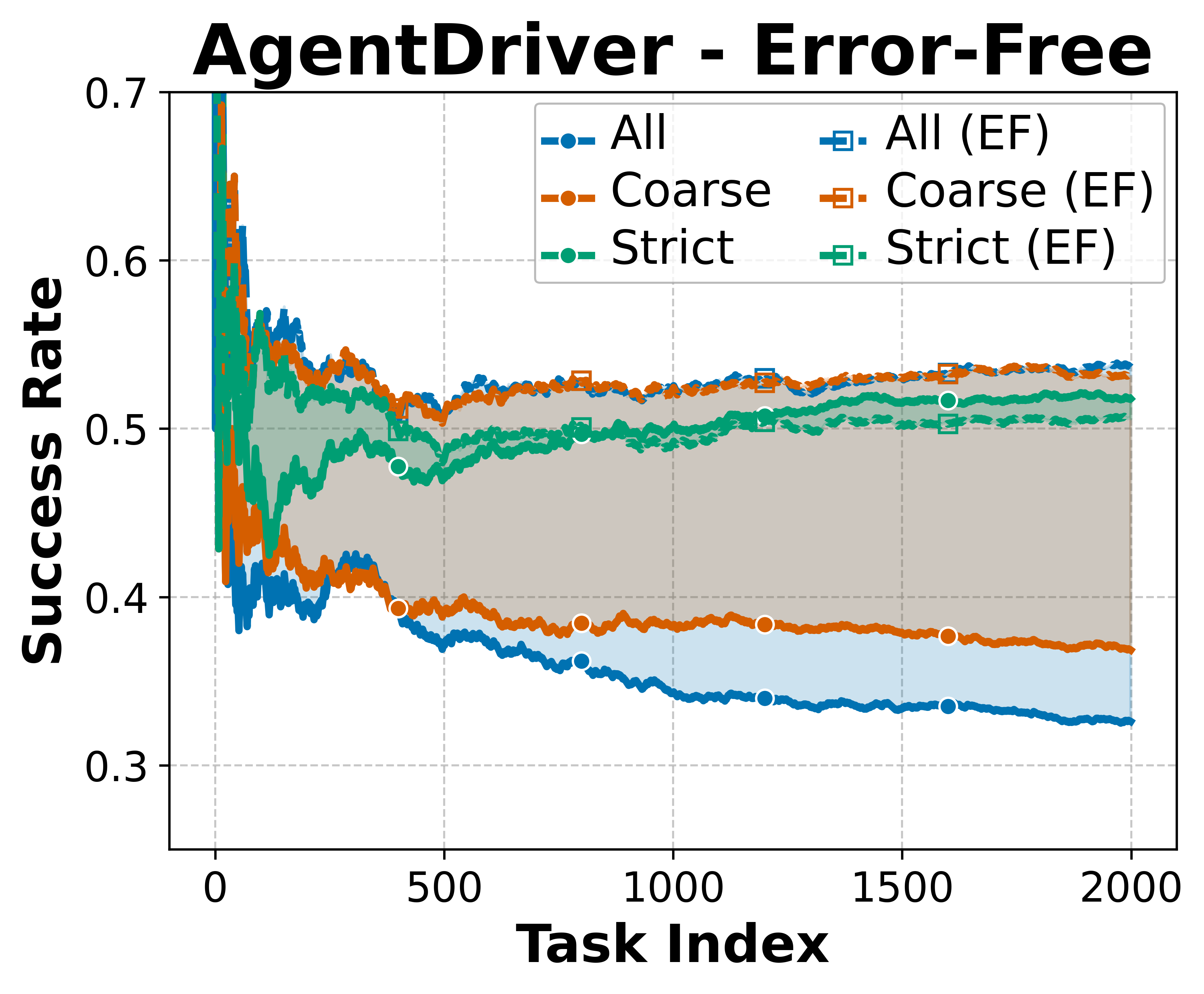}
        \caption{
Comparison of running performance between using the agent output as demonstrations and the error-free (EF) variant using ground-truth. Coarse here uses C1 evaluator.
        }
        \label{fig:error_prop}
    \end{minipage}
\vspace{-0.2in}
\end{figure*}
{
Table~\ref{tab:comparison} summarizes the performance of memory addition strategies guided by five evaluators across four agents.
Although all methods start with the same initial memory, their long-term performances diverge notably. The fixed-memory baseline maintains strong results with RegAgent, AgentDriver, and CIC-IoT Agent—sometimes even surpassing runs guided by coarse evaluators—indicating that noisy or low-quality additions can harm memory utility. In contrast, using a strict evaluator that selectively expands memory with high-quality records consistently yields superior performance, showing that both memory quality and capacity are key to effective long-term learning.

Furthermore, performance relies on the judge's capability for coarse evaluators. For the three model judged agents, fine-tuning the evaluator on only 300 trajectories (C3) already achieves strong long-term improvements, outperforming other coarse evaluators and unfiltered additions. 

As shown in Figure~\ref{fig:ehr_add_acc} and Figure~\ref{fig:other_add_acc} in Appendix~\ref{app:addition_results}, the add-all and some coarse addition approach exhibits flat or declining success rates, suggesting accumulation of flawed records. The fixed-memory baseline performs better but remains static due to its frozen memory. In contrast, strict selective addition and C3 coarse evaluator continue to improve over time, a trend consistent across different LLM backbones (Appendix~\ref{app:diff_llmbackones}) with strict addition.

These results suggest that careful construction of the evaluator is particularly important, as directly applying a vanilla LLM as a trajectory evaluator~\citep{wang2024agent,pan2024autonomous} may lead to a more severe negative impact than manually crafting a small but high-quality dataset.
}

\subsection{Experience-Following Property}
\label{sec:add_experience_follow}

For each memory addition strategy, we measure both the input similarities and output similarities between each query and the memory records retrieved during its execution.
{The computation of both types of similarities for each query can be found in Appendix~\ref{app:detailed_setup}.}
For each of the input similarities and the output similarities, we compute the cumulative average over the entire stream of test query executions to demonstrate the long-term trend.
As shown in Figure~\ref{fig:ehr_add_sim} for RegAgent and AgentDriver, the fixed-memory baseline produces both low input and output similarities, whereas the addition-based methods exhibit higher output similarity as the input similarity grows.
Similar patterns are observed for the other two agents  (Figure~\ref{fig:other_add_sim} in Appendix~\ref{app:addition_results}) and alternative LLM backbones such as state-of-the art GPT-4o and DeepSeek-V3 (Figure~\ref{fig:ablation_different} in Appendix~\ref{app:diff_llmbackones}).
The exhibition of such a strong correlation between the input and output similarities is referred to as the \textbf{experience-following} property of memory-based LLM agents.

Intuitively, this property reflects that LLM agents tend to imitate retrieved experience more closely when their current queries are similar to past examples.
As the memory expands to include a broader range of experiences, the agent is more likely to retrieve highly similar records for new queries.
{\
For instance, in RegAgent, the agent can make predictions either by directly mimicking the closest demonstration or by internal reasoning to guess the mapping rule based on demonstrations.
When the memory size increases and demonstrations become more similar to the query, the agent shows a near-perfect correlation (Pearson $r \approx 1$) between input and output similarity, indicating strong demonstration following when input similarity is high.
In contrast, under the fixed-memory setting, both input and output similarities remain consistently lower than those achieved with memory addition, suggesting that the agent still relies on reasoning rather than pure imitation, and thus achieves surprisingly high performance compared with those using a coarse evaluator with noisy memory. A similar pattern is also observed in AgentDriver.
} 
Therefore, a careful selection strategy with a stronger evaluator for memory addition can facilitate the self-improvement of the agent through the replication of correct executions of similar tasks.
Conversely, indiscriminate memory addition will easily introduce incorrect executions for the agent to learn from, causing a self-degradation of its performance.

\subsection{Error Propagation in Agent Memory}
\label{sec:erro_prop}

While experience-following enables LLM agents to learn from high-quality examples with a strong evaluator, the inclusion of erroneous memory records or seemingly well-performed trajectories with incorrect intermediate attempts remains inevitable in many cases, even under human inspection of the agent’s execution.

When erroneous and noisy memory records are retrieved as demonstrations, these errors can influence the current task execution. If the current execution is then stored in the memory, the error may propagate to future tasks by affecting their execution.
This \textbf{error propagation} poses significant challenges to memory management, causing a deviation between the agent's performance and the optimal level achievable with perfect memory.

In Figure~\ref{fig:error_prop}, we visualize error propagation on RegAgent and AgentDriver.
For each addition strategy in Section \ref{sec:add_setup}, we compare it with a variant that uses the same retrieved examples for each task but replaces the LLM’s execution with the ground-truth output, ensuring error-free\footnote{For AgentDriver, while these trajectories can still be suboptimal, the ground-truth trajectory resembles relatively ``correct'' task executions.} retrieval.

For both agents, we observe an immediate gap in performance compared to their error-free variant.
Moreover, as the execution continues, both add-all and coarse selective addition 
exacerbate such a performance gap.
However, for AgentDrievr with strict selective addition, despite lagging initially, it gradually approaches the ground-truth baseline and even surprisingly surpasses its performance after roughly 2000 executions. 
These results again highlight the importance of both the quantity and quality of memory, underscoring the necessity of carefully selecting and training trajectory evaluators for stable long-term agent execution.

\section{Deletion of Memory}
\label{sec:deletion}
Memory deletion is often necessary for real systems, as the memory cannot grow indefinitely due to the practical constraint of storage.
In this section, we evaluate three memory deletion approaches: the commonly adopted periodic deletion, the proposed history-based deletion guided by future execution utility, and a hybrid approach that combines both.

\subsection{Setup for Memory Deletion Experiments}
\label{sec:deletion_setup}
We follow Section~\ref{sec:add_setup} for the four agent setups we considered.
For the deletion, let \((q_i, e_i)\) be a query-execution pair where \(1 \le i \le N\) is the index. This pair will be removed from the memory bank if it satisfies a specific criterion \(\phi\). We focus on three memory deletion strategies that can be deployed together with selective memory addition (either coarse (automatic) or strict (human) evaluation):

\textbf{1) Periodical-based Deletion:}
Prior research proposes to use a forgetting rate inspired by human cognition to assign deletion probabilities based on how frequently and how recently a memory record is retrieved~\cite{zhong2024memorybank,hou2024my,wang2024trove}. In this work, we adopt a simplified (thus more widely applicable) strategy using a fixed threshold to determine whether a memory record should be deleted based on its past retrieval frequency during a given period.
Specifically, 
Let \(\text{fr}_{t}(q_i, e_i)\) denote the total retrievals of \((q_i, e_i)\) at the current timestamp \(t\), and \(\text{fr}_{t'}(q_i, e_i)\) be the retrievals at an earlier timestamp \(t'\). Define \(\alpha\) as the target retrieval count within \([t', t]\). A memory record will be deleted if \(\phi_{\text{per}}(q_i, e_i, t, t') = 1\) where
\[
\phi_{\text{per}}(q_i, e_i, t, t') 
= \mathbf{1}\!\left[\, 
\operatorname{fr}_{t}(q_i, e_i) - \operatorname{fr}_{t'}(q_i, e_i) 
\le \alpha 
\,\right].
\]
This approach keeps the memory size $M$ bounded {by $M \le\alpha(t - t') K$, where $K$ is the number of retrieved memory per execution}.

\textbf{2) History-Based Deletion:}
{
Following the previous discussion on experience-following and error propagation, we hypothesize a correlation between the quality of experiences stored in memory and the downstream execution quality of future tasks when these experiences are retrieved.}
Thus, we propose a simple history-based deletion strategy guided by the 
utility of stored memory records over time. 
Specifically, suppose \(\Phi\) is a utility evaluator, which could be the same one used for selective addition.
For any timestamp $t$, a record will be removed if a) it has been retrieved for at least $n$ times, and b) the average utility across all its past retrievals is below a prescribed threshold \(\beta\):

\[
\phi_{\text{hist}}(q_i, e_i, t) =
\begin{cases}
\delta(q_i, e_i, t), & \text{if } \text{fr}_t(q_i, e_i) > n, \\
0, & \text{otherwise}.
\end{cases}
\]
where
\[
\delta(q_i, e_i, t) 
= \mathbf{1}\!\left[
  \tfrac{1}{\operatorname{fr}_{t}(q_i, e_i)}
  \sum_{m=1}^{\operatorname{fr}_{t}(q_i, e_i)} 
  \Phi(q_m, e_m)
  \;\le\; \beta
\right].
\]
Here, $\Phi(q_m, e_m)$ denotes the utility obtained from the $m$-th retrieval of the specific memory record $(q_i, e_i)$.
Note that we require a memory record to be retrieved at least $n$ times before being considered for removal to reduce the estimation bias for the average utility.

\textbf{3) Combined Deletion:}
Periodical- and history-based methods can be applied together to jointly enhance the agent's performance and reduce the memory size:
\[
\phi_{\text{comb}}(q_i,e_i,t,t') = 
\phi_{\text{per}}(q_i,e_i,t,t') \,\lor\, \phi_{\text{hist}}(q_i,e_i,t).
\]

\subsection{Strategic Memory Deletion Improves the Agent Performance}
\label{sec:deletion_performance}
\begin{table}[t!]
\caption{Performance of periodical-based deletion, history-based deletion, and the combined approach, when deployed with the two selective addition strategies (strict and coarse with C1 evaluator) and using the same evaluator for history-based and combined deletion.
The history-based approach in real agents often yields the best performance when paired with a strict evaluator among deletion approaches, while the combined method reduces the memory size the most. 
}
\begin{adjustbox}{width=\linewidth} %
\begin{tabular}{c|cc|cc|cc|cc}
\toprule
 \multirow{2}{*}{\textbf{Strategy}} 
& \multicolumn{2}{c|}{\textbf{RegAgents}} 
& \multicolumn{2}{c|}{\textbf{EHRAgents}} 
& \multicolumn{2}{c|}{\textbf{AgentDriver}} 
& \multicolumn{2}{c}{\textbf{CIC-IoT Agent}} \\  
& \textbf{SR. $\uparrow$}  & \textbf{Mem Size $\downarrow$}  
 & \textbf{ACC. $\uparrow$}  & \textbf{Mem Size $\downarrow$} 
& \textbf{SR. $\uparrow$}  & \textbf{Mem Size $\downarrow$}     
& \textbf{ACC. $\uparrow$} & \textbf{Mem Size $\downarrow$} \\ \hline
  \rowcolor{gray!20}\multicolumn{9}{c}{\textbf{Coarse evaluator (C1)}}  \\
\textbf{No del} 
&63.18 & 3511
& 25.91 & 1447 & 36.92 & 1161 & 74.00 & 1030  \\ \hline
\textbf{Period}              
& 60.88&1012 &26.65 &338  & 36.38 & 426 &78.10 & 355 \\ \hline

 \textbf{History}                  
&62.10& 3205& 33.55 & 1004  & 34.00 & 1019  &73.70 & 952\\ \hline 
 \textbf{Combined}
&59.32 &951 &31.47 &279 & 35.62 & {372} &68.80 & 352
\\\hline
  \rowcolor{gray!20}\multicolumn{9}{c}{\textbf{Strict evaluator}}  \\
 \textbf{No del}                 
& \textbf{70.95} & 2938 &38.67 &  1012 & 51.00  & 1178  &85.40 & 904  \\\hline

 \textbf{Period}              
&67.65 &949 &38.59 &302  & 50.94 & 467  &80.80 & 310 \\ \hline

 \textbf{History}                  
& 69.80&2286 &42.06 &784   & \textbf{51.81} & 846 & \textbf{89.60}& 788\\ \hline
 \textbf{Combined}
& 66.58 &\textbf{890} &\textbf{42.34} &\textbf{248}   &49.97  & \textbf{323} &85.50& \textbf{188}
\\
 \bottomrule
\end{tabular}
\end{adjustbox}
\label{tab:deletion}
\vspace{-0.2in}
\end{table}
\begin{figure}[t!]
    \centering
     \includegraphics[width=0.46\linewidth]{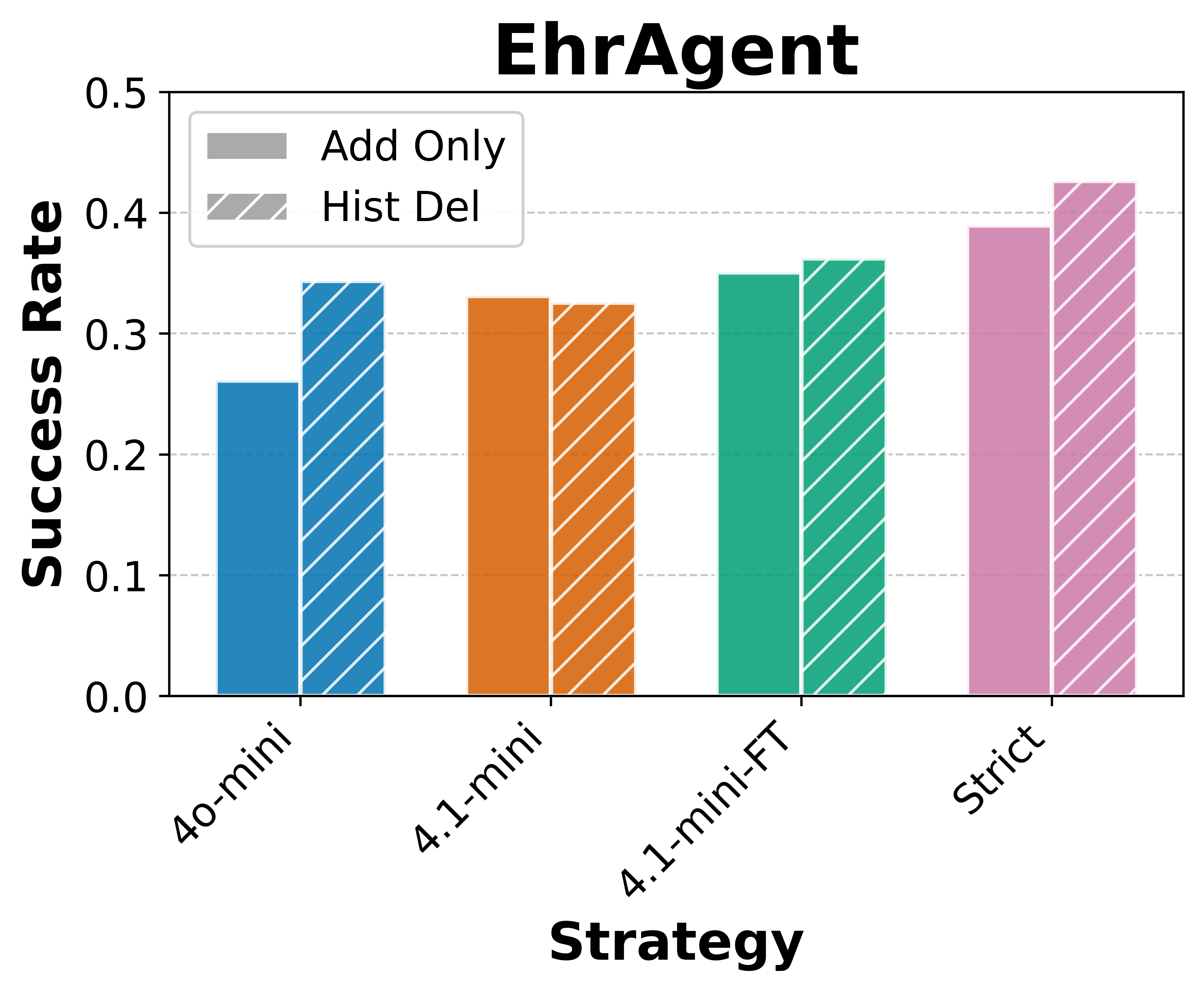} 
        \includegraphics[width=0.46\linewidth]{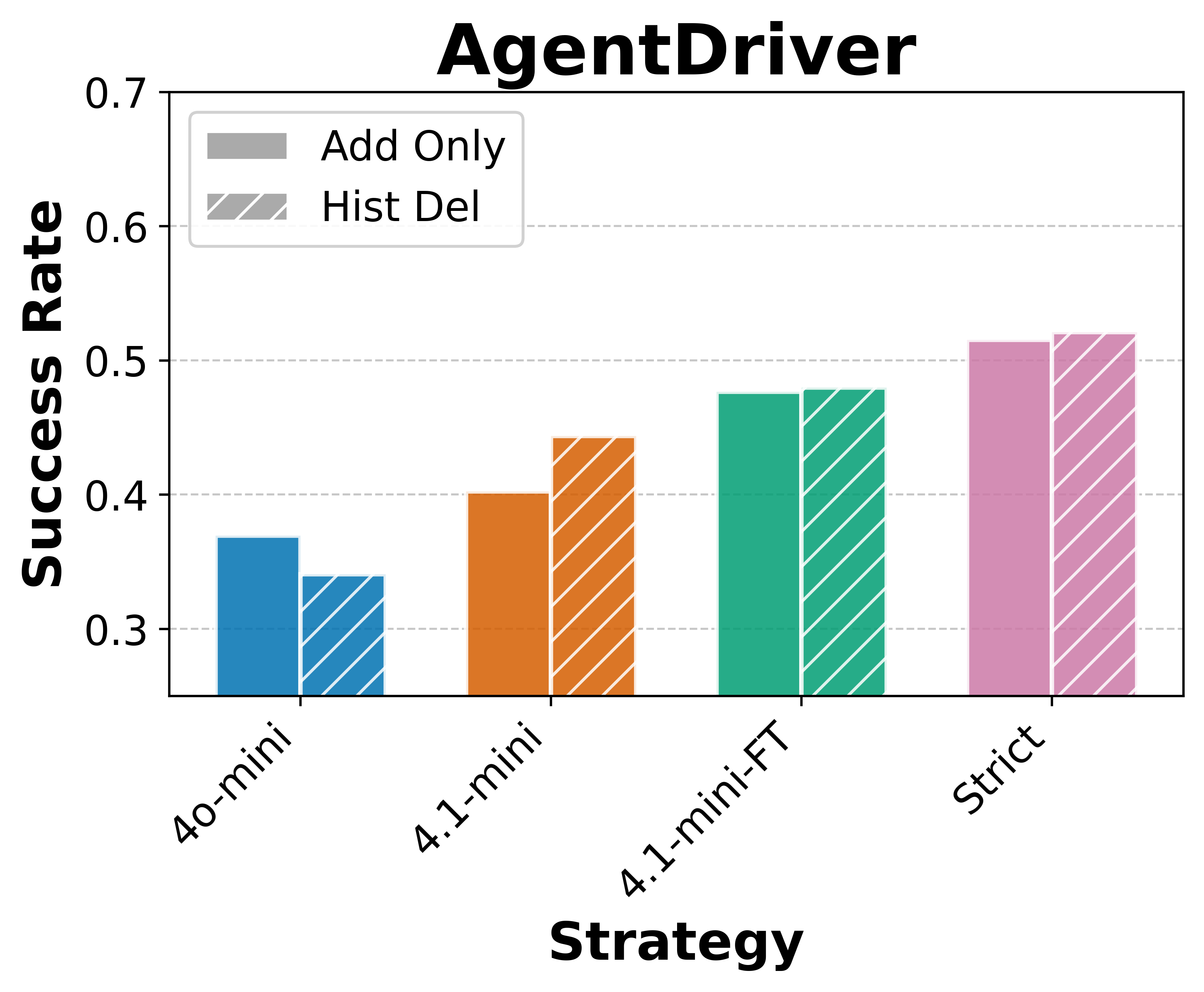}
    \caption{Performance comparison after applying history-based deletion with different evaluators.
    }
    \label{fig:coarse_hist_dif}
     \vspace{-0.2in}
\end{figure}

Table~\ref{tab:deletion} summarizes the effects of three deletion strategies on both agent performance and memory size. 

We first observe that \textit{periodical-based deletion} achieves substantial memory reduction with small performance degradation in general,
indicating that addition-only memory designs often accumulate redundant entries. 

\textit{History-based deletion} and \textit{combined deletion} show more variable results depending on the reliability of the utility evaluator used during addition and deletion. Specifically, when a strict evaluator is employed, history-based deletion leads to notable performance improvements with the non-synthetic agent we considered. 
In addition, for our synthetic {RegAgent}, which exhibits strong experience-following behavior and relies on a large memory capacity, we observe that history-based deletion causes only a small performance drop compared with other strategies, even after removing several hundred frequent executions.

To further understand these improvements, we extend the methodology used in Section~\ref{sec:erro_prop} to construct an error-free memory baseline during agent execution with AgentDriver and RegAgent. As presented in Appendix~\ref{appendix:error_free}, in AgentDriver, history-based deletion with a strict evaluator can even outperform this error-free counterpart with a larger magnitude compared with no deletion. 
This result suggests that selectively retaining experiences with high utility scores on downstream tasks can improve long-term performance.

{
However, when the evaluator used for both addition and deletion is coarse or noisy, history-based and combined deletion can influence the agent’s long-term behavior in different ways.
As shown in Figure~\ref{fig:coarse_hist_dif} and Figure~\ref{fig:appendix_coarse_hist_dif} (Appendix~\ref{appendix:hist-based-evaluators}), the impact of history-based deletion varies notably across evaluators based on LLMs without fine-tuning.
For instance, when using GPT-4o-mini as the history-based evaluator, the deletion strategy yields a clear performance gain on {EhrAgent} but leads to degraded performance on {AgentDriver}.
In contrast, fine-tuned evaluators tend to produce more stable results, achieving performance comparable to their addition-only counterparts while maintaining smaller memory sizes.
}

In summary, our findings show that with a reliable utility evaluator, history-based deletion can improve agent performance.
Additionally, the combined deletion approach offers a strong balance between maintaining performance and reducing memory size, making it a practical and efficient memory management strategy for long-term LLM agent deployment.

\subsection{Misaligned Experience Replay}
\label{sec:erratic_exp}

\begin{figure}
    \centering
     \includegraphics[width=0.46\linewidth]{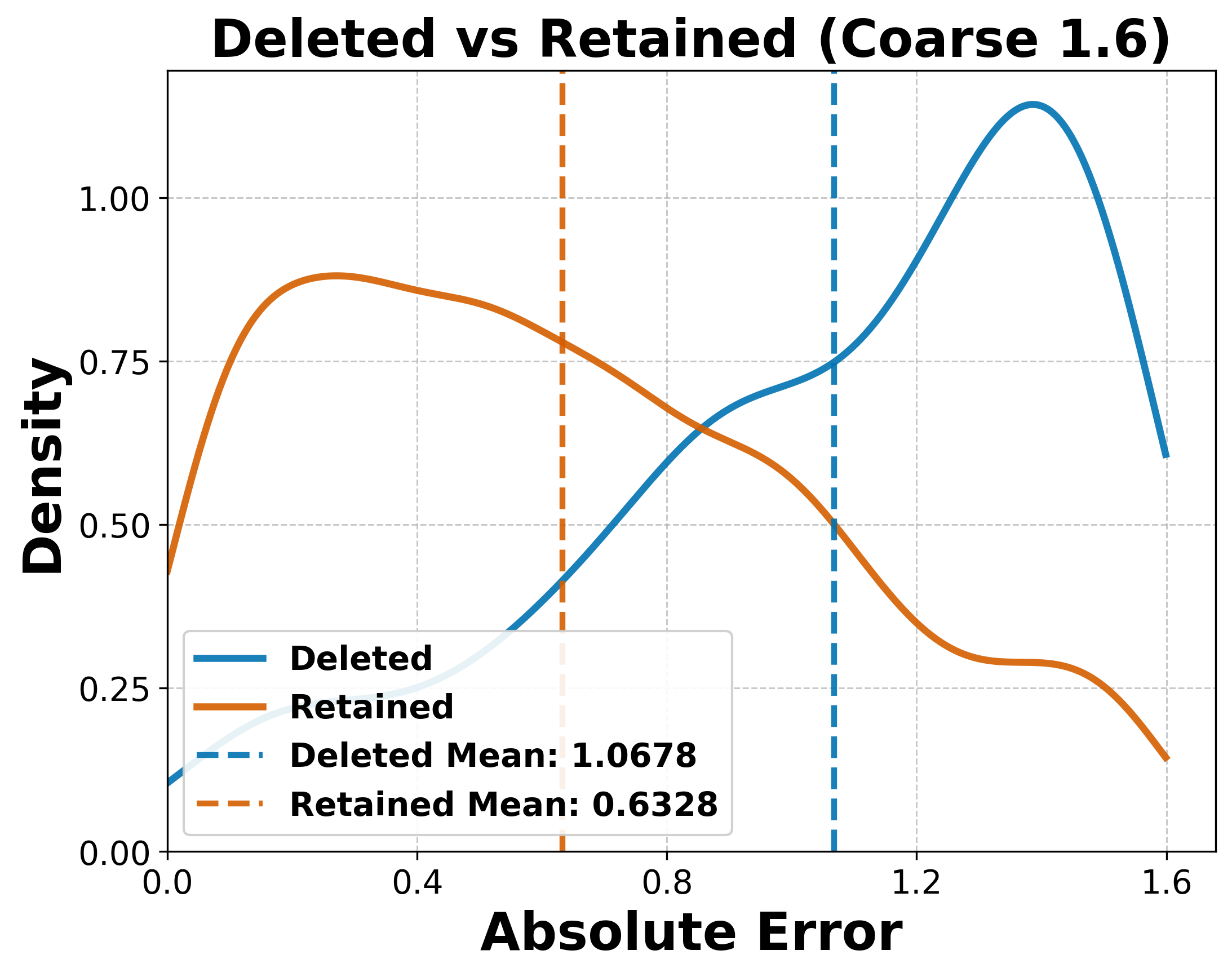} 
        \includegraphics[width=0.46\linewidth]{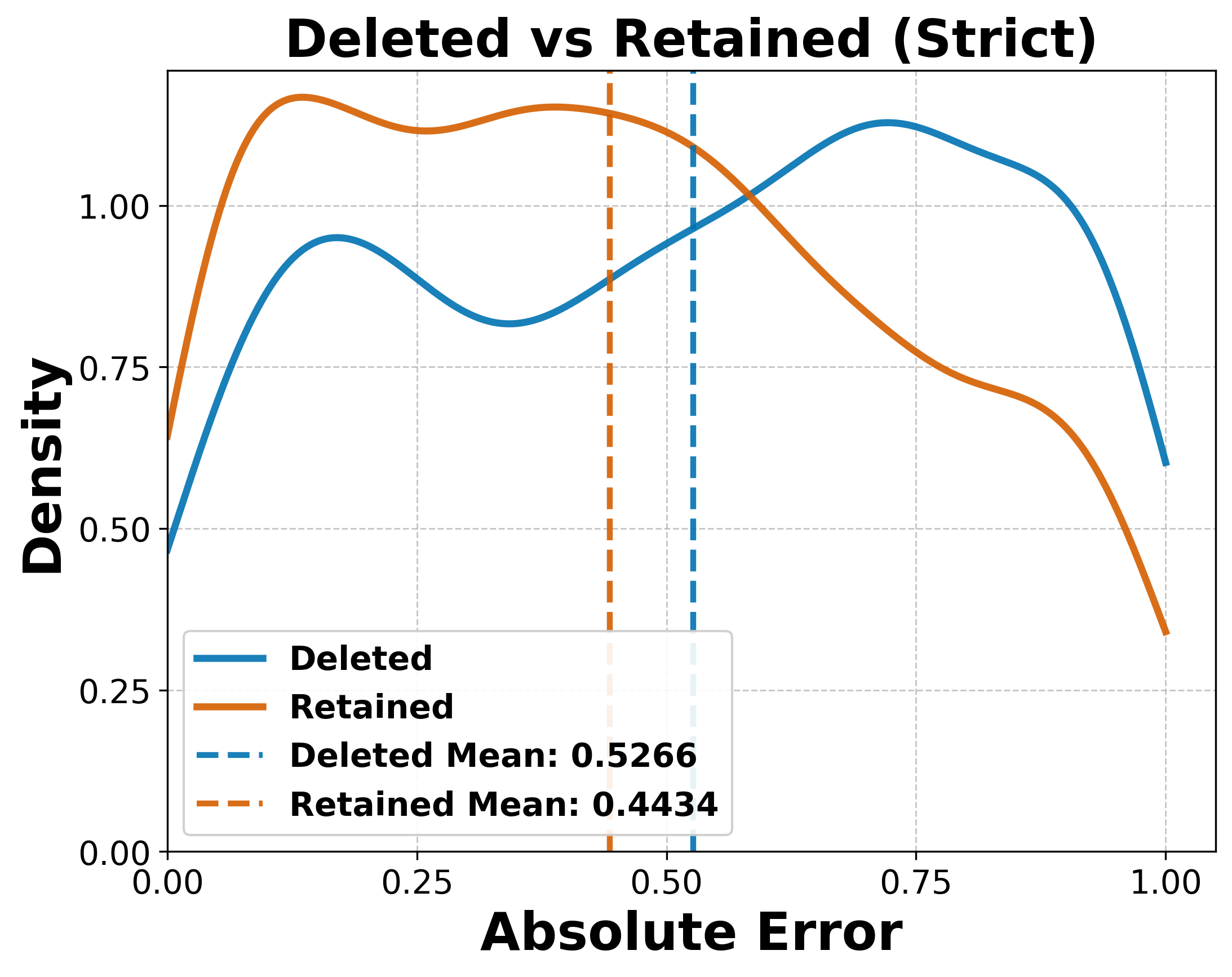}
    \caption{ \textbf{Left:} KDE curve over the absolute error of deleted and retained memory (with retrieved more than 5 times) with C1 evaluator using history-based deletion. \textbf{Right:} KDE curve over the absolute error with strict evaluator.
    }
    \label{fig:demo_quality_diff}
     \vspace{-0.2in}
\end{figure}

{To explain the performance gains achieved by history-based deletion, we hypothesize that certain memory records provide limited or even harmful guidance during task execution.
{
These records, although initially passing the evaluator’s quality filter, may still be misaligned with the objectives of the current task distribution. Such misalignment can arise from inconsistencies between their stored trajectories and the current execution context, or from errors introduced by the evaluator’s limitations. Conditioning on such demonstrations can therefore degrade the quality of task execution.
}
We refer to this phenomenon as \textbf{misaligned experience replay}, emphasizing the \textit{misalignment between the current task and its retrieved demonstrations from the memory bank}.
We attribute the effectiveness of history-based deletion to its ability to remove these misaligned memory records.

{
Since \textit{inconsistencies} between context and task distributions are often case-specific and difficult to capture directly, we instead use the \textit{error} induced by the evaluator’s limitations for clearer illustration, as both factors jointly contribute to misaligned experience replay.
Taking {RegAgent} as an example, the experience quality in terms of the error in the memory record can be directly observed through the discrepancy between its predicted and ground-truth value.
As shown in Figure~\ref{fig:demo_quality_diff}, we plot the kernel density estimation (KDE) curves for both deleted and retained records based on two different evaluators, respectively,
revealing a clear quality gap: retained demonstrations exhibit lower error scores than deleted ones.
Even under strict addition—where only data with error $\le 1$ were initially stored—conditioning on noisier experiences (e.g., entries with error $> 0.5$) can still propagate errors to subsequent tasks more easily, leading to degraded predictions and thus removed by history-based deletion.
Consistent trends across other agents (Appendix~\ref{app:delete_retain_comparison}) further demonstrate that history-based deletion effectively identifies and removes noisy or misaligned memory records with various evaluators.

These findings suggest that experience quality, especially the intrinsic quality of experience itself, is tightly linked to downstream execution quality. 
By effectively removing those demonstrations with low future execution quality through an evaluator, we can mitigate misaligned experience replay.
}

\section{Memory Management under Challenging Scenarios}
\label{sec:challenge}

\subsection{Memory Management with Task distribution shift}  
\label{sec:task_dist}

Task distribution shift occurs when the predominant task type and the expected execution change over time. To simulate a task distribution shift, we construct a modified dataset from the original test sets of EHRAgent and AgentDriver by aggregating their input vectors and reordering them to alter the underlying task distribution. More details for setup can be found in Appendix~\ref{app:task_dist_shift}.

\paragraph{Results}
\begin{figure}[t!]
    \centering
    \includegraphics[width=0.23\textwidth]{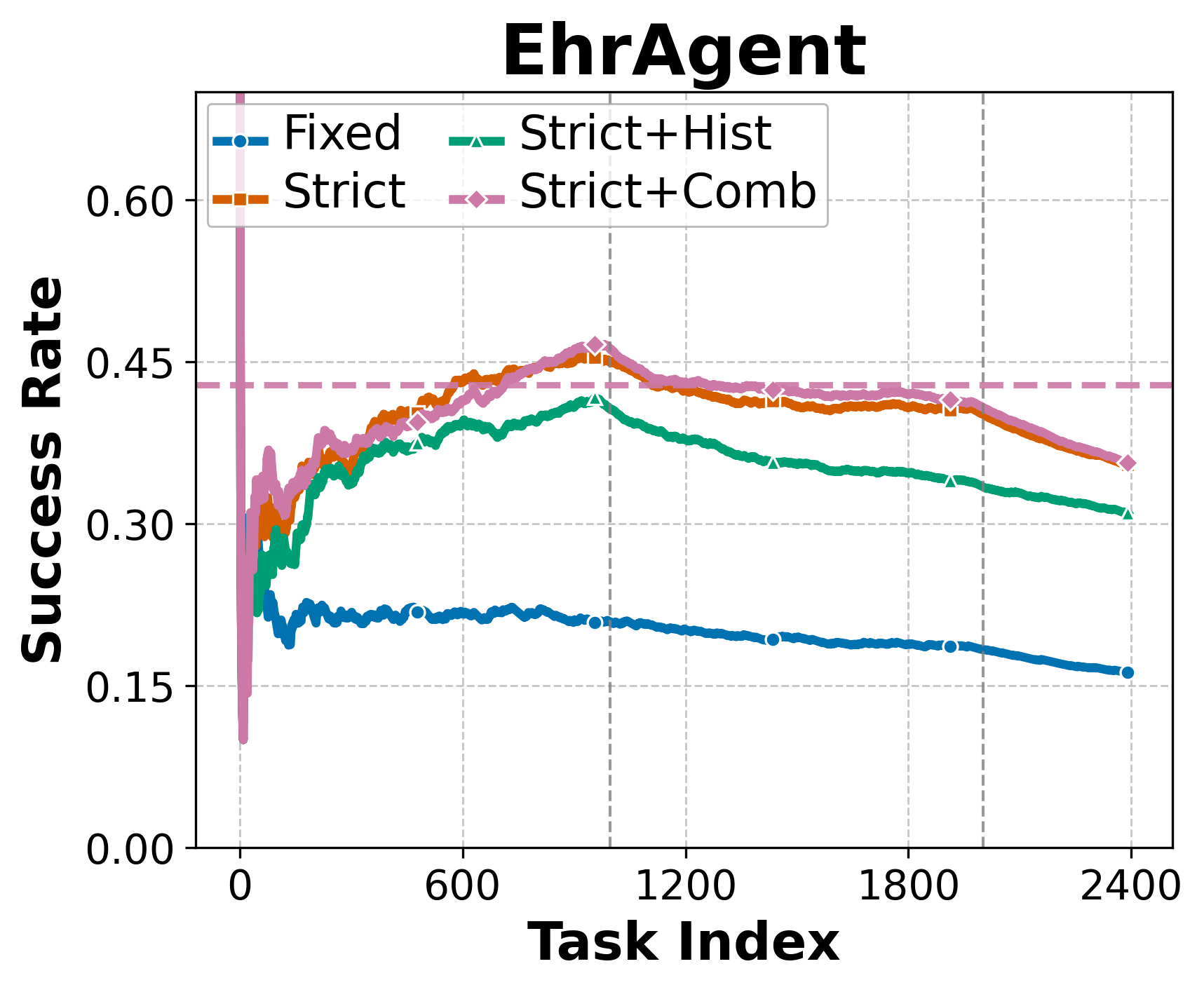}
          \includegraphics[width=0.23\textwidth]{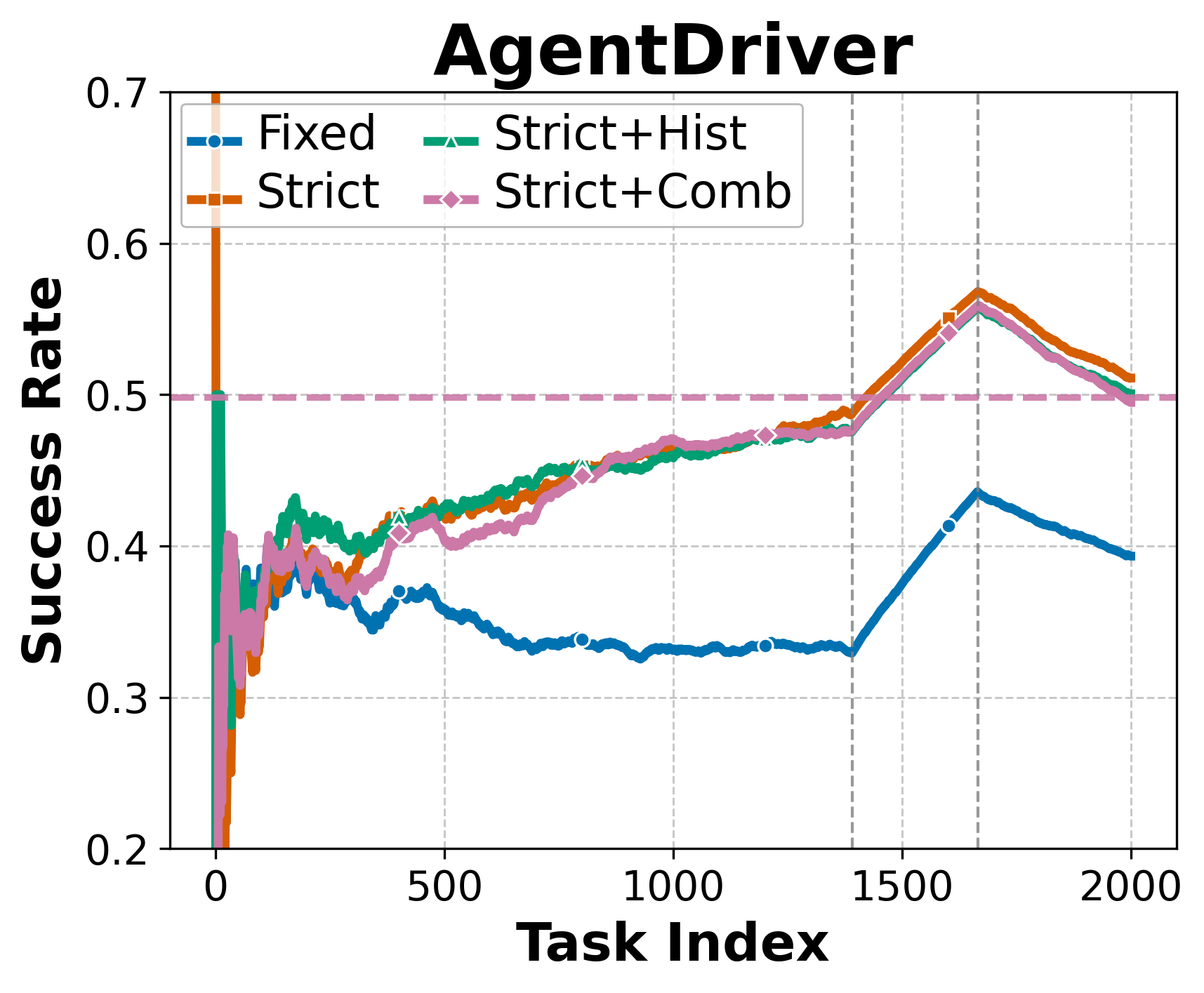} 
    \caption{
Performance comparison under task distribution shift for EHRAgent and AgentDriver.
The vertical line indicates the point at which the task distribution shifts. The horizontal dashed line shows the performance of the combined deletion variant without distribution shift. 
    }
    \label{fig:ehr_distribution_shift}
\vspace{-0.2in}
\end{figure}
In Figure~\ref{fig:ehr_distribution_shift}, we compare performance trends across several memory configurations on EHRAgent and AgentDriver: fixed memory, strict addition, history-based deletion with a strict evaluator, and combined deletion with a strict evaluator. We also include results from a variant of combined deletion (with a strict evaluator) executed under a setting without task distribution shift, as shown in Section~\ref{sec:deletion}.
We observe varying performance dynamics across task distributions. However, in general, the performance gap relative to the no-shift variant remains small. Notably, in AgentDriver, strict addition alone achieves performance that even surpasses the no-shift variant. In contrast, on EHRAgent, history-based deletion underperforms compared to combined deletion. This suggests that in practical scenarios involving distribution shift, periodic deletion—despite its simplicity—can contribute to stabilizing performance.

\subsection{Memory Management with Resource Constraints}
\label{sec:limited_resources}

We also investigate a scenario where the memory capacity is fixed, for example, to its initial size of 100 records for EHRAgent and 180 records for AgentDriver. 
Under this constraint, we modify the combined deletion policy so that after each task execution, it first performs periodical-based deletion and then removes only the record with the least average utility (rather than all records below the utility threshold) if the memory still exceeds the limit after additions.

\paragraph{Results} As shown in Figure~\ref{fig:agentdriver_ciciot} in Appendix~\ref{app:differ_size}, the memory management policies achieve high performance under strict capacity constraints compared with the fixed-memory variant.
By selectively retaining only the most relevant and high-quality records, the agent utilizes limited storage efficiently. 
These results indicate that effective memory management choices can still improve the long-term agent performance in resource-limited settings. 
We also study the relationship between the size of the constrained memory and the performance on AgentDriver, as shown in Figure~\ref{fig:agentdriver_different_size} in Appendix~\ref{app:differ_size}, which demonstrates a gradually converging performance when using a strict utility evaluator.       
This also suggests that naive, unbounded memory growth is unnecessary.

\section{Conclusion}
This paper studies memory management in LLM agents through addition and deletion.
We identify the experience-following phenomenon and reveal two key challenges—error propagation and misaligned experience replay.
Experiments demonstrate that evaluator reliability is critical and that incorporating evaluator signals is essential for effective memory management.

\section*{Limitations}
One limitation of this work lies in the scope of memory management considered.
To build a generalizable understanding of memory dynamics over time, this study focuses on two fundamental operations—memory addition and memory deletion—while omitting more complex and less generalizable mechanisms such as structural transformation~\cite{zeng2024structural}, merging~\cite{liu2023think}, summarization~\cite{pan2025memory}, and reflection~\cite{shinn2024reflexion}.
Consequently, when extending our findings to systems that integrate these advanced updating mechanisms or operate with different agent architectures, additional fine-grained analyses will be helpful.

Another limitation concerns the lack of theoretical guarantees.
Our findings are primarily based on extensive empirical analyses, which, although comprehensive, do not provide formal theoretical proofs.
Given the inherent complexity of agentic systems, deriving such guarantees is often infeasible.
Nonetheless, we believe that the proposed framework offers a solid foundation for future theoretical investigation, particularly through controlled experimentation in our synthetic {RegAgent} environment.

\bibliography{main}

\begin{thebibliography}{35}
\providecommand{\natexlab}[1]{#1}

\bibitem[{Anokhin et~al.(2024)Anokhin, Semenov, Sorokin, Evseev, Burtsev, and Burnaev}]{anokhin2024arigraph}
Petr Anokhin, Nikita Semenov, Artyom Sorokin, Dmitry Evseev, Mikhail Burtsev, and Evgeny Burnaev. 2024.
\newblock Arigraph: Learning knowledge graph world models with episodic memory for llm agents.
\newblock \emph{arXiv preprint arXiv:2407.04363}.

\bibitem[{Beaunieux et~al.(2006)Beaunieux, Hubert, Witkowski, Pitel, Rossi, Danion, Desgranges, and Eustache}]{proceduralmemory}
H~Beaunieux, V~Hubert, T~Witkowski, AL~Pitel, S~Rossi, JM~Danion, B~Desgranges, and F~Eustache. 2006.
\newblock Which processes are involved in cognitive procedural learning?
\newblock \emph{Memory. 2006 Jul;14(5):521-39. doi: 10.1080/09658210500477766. PMID: 16754239.}

\bibitem[{Caesar et~al.(2020)Caesar, Bankiti, Lang, Vora, Liong, Xu, Krishnan, Pan, Baldan, and Beijbom}]{caesar2020nuscenes}
Holger Caesar, Varun Bankiti, Alex~H Lang, Sourabh Vora, Venice~Erin Liong, Qiang Xu, Anush Krishnan, Yu~Pan, Giancarlo Baldan, and Oscar Beijbom. 2020.
\newblock nuscenes: A multimodal dataset for autonomous driving.
\newblock In \emph{Proceedings of the IEEE/CVF conference on computer vision and pattern recognition}, pages 11621--11631.

\bibitem[{Hou et~al.(2024)Hou, Tamoto, and Miyashita}]{hou2024my}
Yuki Hou, Haruki Tamoto, and Homei Miyashita. 2024.
\newblock " my agent understands me better": Integrating dynamic human-like memory recall and consolidation in llm-based agents.
\newblock In \emph{Extended Abstracts of the CHI Conference on Human Factors in Computing Systems}, pages 1--7.

\bibitem[{Hu et~al.(2024)Hu, Chen, Chen, Mu, Shao, and Luo}]{hu2024hiagent}
Mengkang Hu, Tianxing Chen, Qiguang Chen, Yao Mu, Wenqi Shao, and Ping Luo. 2024.
\newblock Hiagent: Hierarchical working memory management for solving long-horizon agent tasks with large language model.
\newblock \emph{arXiv preprint arXiv:2408.09559}.

\bibitem[{{Johnson} et~al.(2016){Johnson}, {Pollard}, {Shen}, {Lehman}, {Feng}, {Ghassemi}, {Moody}, {Szolovits}, {Anthony Celi}, and {Mark}}]{2016NatSD...360035J}
Alistair E.~W. {Johnson}, Tom~J. {Pollard}, Lu~{Shen}, Li-Wei~H. {Lehman}, Mengling {Feng}, Mohammad {Ghassemi}, Benjamin {Moody}, Peter {Szolovits}, Leo {Anthony Celi}, and Roger~G. {Mark}. 2016.
\newblock \href {https://doi.org/10.1038/sdata.2016.35} {{MIMIC-III, a freely accessible critical care database}}.
\newblock \emph{Scientific Data}, 3:160035.

\bibitem[{Kumar(2020)}]{Kumar2020SemanticMA}
Abhilasha~Ashok Kumar. 2020.
\newblock \href {https://api.semanticscholar.org/CorpusID:221495897} {Semantic memory: A review of methods, models, and current challenges}.
\newblock \emph{Psychonomic Bulletin \& Review}, 28:40 -- 80.

\bibitem[{Kusupati et~al.(2022)Kusupati, Bhatt, Rege, Wallingford, Sinha, Ramanujan, Howard-Snyder, Chen, Kakade, Jain et~al.}]{kusupati2022matryoshka}
Aditya Kusupati, Gantavya Bhatt, Aniket Rege, Matthew Wallingford, Aditya Sinha, Vivek Ramanujan, William Howard-Snyder, Kaifeng Chen, Sham Kakade, Prateek Jain, and 1 others. 2022.
\newblock Matryoshka representation learning.
\newblock \emph{Advances in Neural Information Processing Systems}, 35:30233--30249.

\bibitem[{Lampinen et~al.(2025)Lampinen, Engelcke, Li, Chaudhry, and McClelland}]{lampinen2025latentlearningepisodicmemory}
Andrew~Kyle Lampinen, Martin Engelcke, Yuxuan Li, Arslan Chaudhry, and James~L. McClelland. 2025.
\newblock \href {https://arxiv.org/abs/2509.16189} {Latent learning: episodic memory complements parametric learning by enabling flexible reuse of experiences}.

\bibitem[{Li et~al.(2023)Li, Yu, Li, Chen, and Khashanah}]{li2023tradinggpt}
Yang Li, Yangyang Yu, Haohang Li, Zhi Chen, and Khaldoun Khashanah. 2023.
\newblock Tradinggpt: Multi-agent system with layered memory and distinct characters for enhanced financial trading performance.
\newblock \emph{arXiv preprint arXiv:2309.03736}.

\bibitem[{Liang et~al.(2024)Liang, He, Xia, Song, Wang, Tao, Sun, Yuan, Su, Li et~al.}]{liang2024self}
Xuechen Liang, Yangfan He, Yinghui Xia, Xinyuan Song, Jianhui Wang, Meiling Tao, Li~Sun, Xinhang Yuan, Jiayi Su, Keqin Li, and 1 others. 2024.
\newblock Self-evolving agents with reflective and memory-augmented abilities.
\newblock \emph{arXiv preprint arXiv:2409.00872}.

\bibitem[{Liu et~al.(2023)Liu, Yang, Shen, Hu, Zhang, Gu, and Zhang}]{liu2023think}
Lei Liu, Xiaoyan Yang, Yue Shen, Binbin Hu, Zhiqiang Zhang, Jinjie Gu, and Guannan Zhang. 2023.
\newblock Think-in-memory: Recalling and post-thinking enable llms with long-term memory.
\newblock \emph{arXiv preprint arXiv:2311.08719}.

\bibitem[{Luo et~al.(2024)Luo, Xu, Liu, Pasupat, and Kazemi}]{luo2024context}
Man Luo, Xin Xu, Yue Liu, Panupong Pasupat, and Mehran Kazemi. 2024.
\newblock In-context learning with retrieved demonstrations for language models: A survey.
\newblock \emph{arXiv preprint arXiv:2401.11624}.

\bibitem[{Mao et~al.(2024)Mao, Ye, Qian, Pavone, and Wang}]{mao2023language}
Jiageng Mao, Junjie Ye, Yuxi Qian, Marco Pavone, and Yue Wang. 2024.
\newblock \href {https://openreview.net/forum?id=UPE6WYE8vg} {A language agent for autonomous driving}.
\newblock In \emph{First Conference on Language Modeling}.

\bibitem[{Neto et~al.(2023)Neto, Dadkhah, Ferreira, Zohourian, Lu, and Ghorbani}]{ciciot}
Euclides Carlos~Pinto Neto, Sajjad Dadkhah, Raphael Ferreira, Alireza Zohourian, Rongxing Lu, and Ali~A. Ghorbani. 2023.
\newblock \href {https://doi.org/10.3390/s23135941} {Ciciot2023: A real-time dataset and benchmark for large-scale attacks in iot environment}.
\newblock \emph{Sensors}, 23(13).

\bibitem[{Nuxoll and Laird(2007)}]{Nuxoll2007ExtendingCA}
Andrew Nuxoll and John~E. Laird. 2007.
\newblock \href {https://api.semanticscholar.org/CorpusID:7897551} {Extending cognitive architecture with episodic memory}.
\newblock In \emph{AAAI Conference on Artificial Intelligence}.

\bibitem[{Pan et~al.(2024)Pan, Zhang, Tomlin, Zhou, Levine, and Suhr}]{pan2024autonomous}
Jiayi Pan, Yichi Zhang, Nicholas Tomlin, Yifei Zhou, Sergey Levine, and Alane Suhr. 2024.
\newblock Autonomous evaluation and refinement of digital agents.
\newblock \emph{arXiv preprint arXiv:2404.06474}.

\bibitem[{Pan et~al.(2025)Pan, Wu, Jiang, Luo, Cheng, Li, Yang, Lin, Zhao, Qiu et~al.}]{pan2025memory}
Zhuoshi Pan, Qianhui Wu, Huiqiang Jiang, Xufang Luo, Hao Cheng, Dongsheng Li, Yuqing Yang, Chin-Yew Lin, H~Vicky Zhao, Lili Qiu, and 1 others. 2025.
\newblock On memory construction and retrieval for personalized conversational agents.
\newblock \emph{arXiv preprint arXiv:2502.05589}.

\bibitem[{Park et~al.(2023)Park, O'Brien, Cai, Morris, Liang, and Bernstein}]{park2023generative}
Joon~Sung Park, Joseph O'Brien, Carrie~Jun Cai, Meredith~Ringel Morris, Percy Liang, and Michael~S Bernstein. 2023.
\newblock Generative agents: Interactive simulacra of human behavior.
\newblock In \emph{Proceedings of the 36th annual acm symposium on user interface software and technology}, pages 1--22.

\bibitem[{Shi et~al.(2024)Shi, Xu, Zhuang, Yu, Zhang, Wu, Zhu, Ho, Yang, and Wang}]{shi2024ehragent}
Wenqi Shi, Ran Xu, Yuchen Zhuang, Yue Yu, Jieyu Zhang, Hang Wu, Yuanda Zhu, Joyce Ho, Carl Yang, and May~D Wang. 2024.
\newblock Ehragent: Code empowers large language models for complex tabular reasoning on electronic health records.
\newblock \emph{arXiv preprint arXiv:2401.07128}.

\bibitem[{Shinn et~al.(2024)Shinn, Cassano, Gopinath, Narasimhan, and Yao}]{shinn2024reflexion}
Noah Shinn, Federico Cassano, Ashwin Gopinath, Karthik Narasimhan, and Shunyu Yao. 2024.
\newblock Reflexion: Language agents with verbal reinforcement learning.
\newblock \emph{Advances in Neural Information Processing Systems}, 36.

\bibitem[{Sumers et~al.(2023)Sumers, Yao, Narasimhan, and Griffiths}]{sumers2023cognitive}
Theodore~R Sumers, Shunyu Yao, Karthik Narasimhan, and Thomas~L Griffiths. 2023.
\newblock Cognitive architectures for language agents.
\newblock \emph{arXiv preprint arXiv:2309.02427}.

\bibitem[{Wang et~al.(2023)Wang, Xie, Jiang, Mandlekar, Xiao, Zhu, Fan, and Anandkumar}]{wang2023voyager}
Guanzhi Wang, Yuqi Xie, Yunfan Jiang, Ajay Mandlekar, Chaowei Xiao, Yuke Zhu, Linxi Fan, and Anima Anandkumar. 2023.
\newblock Voyager: An open-ended embodied agent with large language models.
\newblock \emph{arXiv preprint arXiv:2305.16291}.

\bibitem[{Wang et~al.(2024{\natexlab{a}})Wang, Ma, Feng, Zhang, Yang, Zhang, Chen, Tang, Chen, Lin et~al.}]{wang2024survey}
Lei Wang, Chen Ma, Xueyang Feng, Zeyu Zhang, Hao Yang, Jingsen Zhang, Zhiyuan Chen, Jiakai Tang, Xu~Chen, Yankai Lin, and 1 others. 2024{\natexlab{a}}.
\newblock A survey on large language model based autonomous agents.
\newblock \emph{Frontiers of Computer Science}, 18(6):186345.

\bibitem[{Wang et~al.(2024{\natexlab{b}})Wang, Fried, and Neubig}]{wang2024trove}
Zhiruo Wang, Daniel Fried, and Graham Neubig. 2024{\natexlab{b}}.
\newblock Trove: Inducing verifiable and efficient toolboxes for solving programmatic tasks.
\newblock \emph{arXiv preprint arXiv:2401.12869}.

\bibitem[{Wang et~al.(2025)Wang, Mao, Fried, and Neubig}]{wang2024agent}
Zora~Zhiruo Wang, Jiayuan Mao, Daniel Fried, and Graham Neubig. 2025.
\newblock \href {https://openreview.net/forum?id=NTAhi2JEEE} {Agent workflow memory}.
\newblock In \emph{Forty-second International Conference on Machine Learning}.

\bibitem[{Xiang et~al.(2024)Xiang, Zheng, Li, Hong, Li, Xie, Zhang, Xiong, Xie, Yang et~al.}]{xiang2024guardagent}
Zhen Xiang, Linzhi Zheng, Yanjie Li, Junyuan Hong, Qinbin Li, Han Xie, Jiawei Zhang, Zidi Xiong, Chulin Xie, Carl Yang, and 1 others. 2024.
\newblock Guardagent: Safeguard llm agents by a guard agent via knowledge-enabled reasoning.
\newblock \emph{arXiv preprint arXiv:2406.09187}.

\bibitem[{Xu et~al.(2025)Xu, Mei, Gao, Tan, Liang, and Zhang}]{xu2025mem}
Wujiang Xu, Kai Mei, Hang Gao, Juntao Tan, Zujie Liang, and Yongfeng Zhang. 2025.
\newblock A-mem: Agentic memory for llm agents.
\newblock \emph{arXiv preprint arXiv:2502.12110}.

\bibitem[{Yin et~al.(2024)Yin, Sun, Guo, Zeng, Cheng, Qiu, and Huang}]{yin2024explicit}
Zhangyue Yin, Qiushi Sun, Qipeng Guo, Zhiyuan Zeng, Qinyuan Cheng, Xipeng Qiu, and Xuan-Jing Huang. 2024.
\newblock Explicit memory learning with expectation maximization.
\newblock In \emph{Proceedings of the 2024 Conference on Empirical Methods in Natural Language Processing}, pages 16618--16635.

\bibitem[{Zeng et~al.(2024)Zeng, Fang, Liu, and Meng}]{zeng2024structural}
Ruihong Zeng, Jinyuan Fang, Siwei Liu, and Zaiqiao Meng. 2024.
\newblock On the structural memory of llm agents.
\newblock \emph{arXiv preprint arXiv:2412.15266}.

\bibitem[{Zhang et~al.(2024)Zhang, Bo, Ma, Li, Chen, Dai, Zhu, Dong, and Wen}]{zhang2024survey}
Zeyu Zhang, Xiaohe Bo, Chen Ma, Rui Li, Xu~Chen, Quanyu Dai, Jieming Zhu, Zhenhua Dong, and Ji-Rong Wen. 2024.
\newblock A survey on the memory mechanism of large language model based agents.
\newblock \emph{arXiv preprint arXiv:2404.13501}.

\bibitem[{Zhao et~al.(2024)Zhao, Huang, Xu, Lin, Liu, and Huang}]{zhao2024expel}
Andrew Zhao, Daniel Huang, Quentin Xu, Matthieu Lin, Yong-Jin Liu, and Gao Huang. 2024.
\newblock Expel: Llm agents are experiential learners.
\newblock In \emph{Proceedings of the AAAI Conference on Artificial Intelligence}, volume~38, pages 19632--19642.

\bibitem[{Zheng et~al.(2024)Zheng, Wang, Wang, and An}]{zheng2023synapse}
Longtao Zheng, Rundong Wang, Xinrun Wang, and Bo~An. 2024.
\newblock \href {https://openreview.net/forum?id=Pc8AU1aF5e} {Synapse: Trajectory-as-exemplar prompting with memory for computer control}.
\newblock In \emph{The Twelfth International Conference on Learning Representations}.

\bibitem[{Zhong et~al.(2024)Zhong, Guo, Gao, Ye, and Wang}]{zhong2024memorybank}
Wanjun Zhong, Lianghong Guo, Qiqi Gao, He~Ye, and Yanlin Wang. 2024.
\newblock Memorybank: Enhancing large language models with long-term memory.
\newblock In \emph{Proceedings of the AAAI Conference on Artificial Intelligence}, volume~38, pages 19724--19731.

\bibitem[{Zhou et~al.(2025)Zhou, Chen, Guo, Yan, Lee, Wang, Lee, Zhang, Shao, Yang et~al.}]{zhou2025agentfly}
Huichi Zhou, Yihang Chen, Siyuan Guo, Xue Yan, Kin~Hei Lee, Zihan Wang, Ka~Yiu Lee, Guchun Zhang, Kun Shao, Linyi Yang, and 1 others. 2025.
\newblock Agentfly: Fine-tuning llm agents without fine-tuning llms.
\newblock \emph{arXiv preprint arXiv:2508.16153}.

\end{thebibliography}

\clearpage
\appendix
\section{Detailed experimental setups.}
\label{app:detailed_setup}
\subsection{Agent details and functionality}
\label{app:detail_functionality}
\begin{table*}[t!]
\small
\centering
\caption{Details of Agents and Their Functionality}
\resizebox{\textwidth}{!}{
\begin{tabular}{c|p{2.8cm}|p{1.8cm}|p{3.2cm}|p{2.2cm}|c}
\toprule
\multirow{2}{*}{\textbf{Agent Name}} & \multirow{2}{*}{\textbf{Task}} & \multirow{2}{*}{\textbf{Input}} & \multirow{2}{*}{\textbf{Output}} & \textbf{Retrieve} & \multirow{2}{*}{\textbf{\#Experiences}} \\ 
&&&&\textbf{Feature}&\\\midrule
RegAgent & Regression prediction & Input vector & Output number& Input vector & 6 \\ \midrule
\multirow{2}{*}{EHRAgents} & \multirow{2}{*}{EHR tasks} & Task & \multirow{2}{*}{Code} & Text & \multirow{2}{*}{4} \\ 
& & text query & &embedding& \\\midrule
\multirow{2}{*}{AgentDriver}  & Autonomous  & Vehicle & \multirow{2}{*}{Predicted trajectory} & Ego state, goal,  & \multirow{2}{*}{1} \\
& Driving & state data & & history trajectory&\\\midrule

\multirow{2}{*}{CIC-IoT Agent} & IoT Traffic Detection  & IoT packet & Reasoning and prediction & \multirow{2}{*}{IoT features} & \multirow{2}{*}{3} \\ 
& (8 categories) & data & attack type &  &  \\\bottomrule
\end{tabular}
}
\label{tab:agent_table}
\end{table*}

\paragraph{RegAgents}
RegAgent serves as a synthetic and controlled environment to study the effects of memory management on long-term agent performance.
We first generate an implicit weight vector $w \in \mathbb{R}^6$, and the task is to predict the correct output $y = w^{\top}x$ by retrieving the six most relevant demonstrations from memory.
Specifically, we generate 100 input vectors $x$ sampled from three Gaussian distributions $N(\mu, 1)$ with $\mu \in {-0.5, 0, 0.5}$, each of six dimensions, and compute $y = w^{\top}x + \epsilon$, where $\epsilon$ is bounded noise within $[-1, 1]$. These pairs form the initial memory bank.
During actual execution, we collect 4000 input–output pairs from the same three distributions. A prediction $\hat{y}$ is considered successful if $|\hat{y} - y| \le 1$, and we use the success rate as the primary performance metric.

For retrieval, input similarity is measured by cosine similarity between input vectors, while output similarity is defined as
\(
\text{output\_similarity} = \exp(-\gamma |\mathbf{x}_1 - \mathbf{x}_2|^2),
\)
where $\gamma = 1.0$.

This task is highly controllable and well-suited for our study, as it (1) relies solely on the retrieved demonstrations, (2) allows the agent to either reason or directly mimic the demonstrations to produce outputs, and (3) enables clear definitions of error and addition criteria.

For evaluation, the strict evaluator applies a threshold of 1.0 on the absolute error between prediction and ground truth. Coarse evaluators use thresholds of 1.6 (C1), 1.4 (C2), and 1.2 (C3), respectively.
Periodic deletion is performed with 500 steps with $\alpha = 0$. History-based deletion is applied with a minimum deletion frequency of 5 and a threshold $\beta = 0.5$ across all setups. The combined deletion strategy reuses all hyperparameters above.
We also list the task prompt in Appendix~\ref{app:regagent}
\paragraph{EHRAgents}
Following EHRAgent~\cite{shi2024ehragent}, our agent is a code-generation system that enables clinicians to interact with electronic health records (EHRs) through natural language queries. We adopt the MIMIC-III dataset~\cite{2016NatSD...360035J} for evaluation. After filtering out duplicated and unanswerable entries, we obtain 2,392 tasks in total. During each test query, the agent retrieves four past experiences from an initial memory bank containing 100 records, based on maximum cosine similarity computed using a general text embedding model. We report task performance in terms of accuracy (ACC).

For our implementation, we use OpenAI text-embedding-3-large as the text encoder.
For the coarse evaluator, we use the built-in LLM-generated termination signal, which reflects if LLM considers the task to be completed. 
For the strict evaluator, we use exact match to determine whether the generated answer is correct.
For input similarity, we take the highest cosine similarity between the text embedding of all retrieved experiences and the task.
For output similarity, we use the pycode\_similar package\footnote{\url{https://github.com/fyrestone/pycode_similar.git}} to detect the code plagiarism score between the execution from the retrieved memory with the highest input similarity and the current model execution. 
For periodic deletion, we use a period of 200 and $\alpha = 0$.
For history-based deletion, we set the minimal deleting frequency to 5 and the threshold $\beta$ to 0.3 for the coarse evaluator using GPT-4o-mini and GPT-4.1-mini, and 0.7 for the strict and GPT-4.1-mini fine-tuned evaluator.
The combined deletion reuses all hyperparameters above.

\paragraph{AgentDriver}
Following AgentDriver~\cite{mao2023language}, we adopt an LLM-based autonomous driving agent that integrates common sense and experiential knowledge into memory records. We use the nuScenes dataset~\cite{caesar2020nuscenes} and randomly sample 2,000 test cases from its test split. The initial memory contains 180 experiences randomly drawn from the nuScenes training set.

For the retrieval strategy, the original retrieval pipeline proposed by~\cite{mao2023language} consists of a two-step process: first, selecting the top-3 experiences based on vector similarity, and then prompting an LLM to choose the most relevant one. To improve reproducibility, we simplify this process to a single-step retrieval, selecting only the top-1 experience based on vector similarity.
For the coarse evaluator, we employ an LLM to assess the quality of retrieved experiences. The LLM outputs either ``yes'' or ``no'' to indicate whether a given experience is a ``good'' experience or a ``bad'' one. The prompt used for this LLM-based evaluation is provided in Appendix~\ref{ap:coarse-evaluator-prompts}.
For the strict evaluator, we measure the UniAD 3-second average L2 distance between the predicted trajectory and the ground truth. Experiences with an L2 error lower than 2.5 are added to the memory bank.
For input similarity, we adhere to the computation method from the original paper. Specifically, we compute the exponential negative L2 distance between the query vector and stored experiences, using the same coefficients as in~\cite{mao2023language}.
For output similarity, we compute the radial basis function (RBF) kernel between the predicted trajectory and the ground-truth trajectory. The similarity score is calculated with the same formula in RegAgent:

\[
\text{output\_similarity} = \exp \left( -\gamma \|\mathbf{v_1} - \mathbf{v_2}\|^2 \right).
\]

In our experiments, we set $\gamma = 1.0$.
For periodic deletion, we apply a deletion period of 500 steps with a threshold of $\alpha = 0$.
For history-based deletion, we set a threshold of 5.0 for the UniAD 3-second average L2 distance. An experience is deleted if it has been retrieved at least 3 times and the mean UniAD 3-second average L2 distance across all retrievals exceeds this threshold. Coarse deletion leverages the predicted success rate from the coarse evaluator and deletes a record if the accumulated success rate is less than 0.5.
The combined deletion mechanism incorporates all the aforementioned hyperparameters.

\paragraph{CIC-IoT}
Following the setup in the CIC-IoT benchmark~\cite{ciciot}, we implement a network traffic detection agent designed to predict traffic types from IoT packet features. The original CIC-IoT dataset contains 34 attack classes. Since our implementation relies solely on single-flow features, we exclude categories that cannot be distinguished based on such features. For example, we retain \textit{DDoS-
HTTP\_Flood} while removing \textit{DoS-
HTTP\_Flood} due to their high similarity under single-flow representations. After filtering, we retain 8 representative classes that remain distinguishable using single-flow features, and construct each test query by formatting these features into a predefined prompt template. We evaluate CIC-IoT Agent on the CIC-IoT dataset, randomly sampling 1,000 test cases from its training split. During each execution, the agent retrieves three past experiences using a feature-based similarity approach. The initial memory bank is initialized with 100 synthetic records generated by GPT-4o-mini from a disjoint training subset. We report task performance in terms of accuracy (ACC).

For the text encoder, we use OpenAI text-embedding-3-large.
For the coarse evaluator, similar to AgentDriver, we employ an LLM to assess the quality of the experiences. The prompt for the CIC-IoT Agent's LLM evaluator is provided in Appendix~\ref{ap:coarse-evaluator-prompts}.
For the strict evaluator, we use string matching to check if the ground-truth is contained in the generated answer. For input similarity, we use a feature-based approach. Specifically, we compute the relative change across all features and take the average.  
For continuous features, the relative change between two inputs, $\mathbf{input}_1$ and $\mathbf{input}_2$, for feature $f_i$ is computed as:
\[
    S_{\text{cont}}(f_i) = \frac{| \mathbf{input}_1(f_i) - \mathbf{input}_2(f_i) |}{\max(|\mathbf{input}_1(f_i)|, |\mathbf{input}_2(f_i)|)}
\]
For discrete features, we define the relative change as follows:
\[
    S_{\text{disc}}(f_i) =
    \begin{cases}
        0, & \text{if } \mathbf{input}_1(f_i) = \mathbf{input}_2(f_i) \\
        1, & \text{otherwise}
    \end{cases}
\]
For output similarity, we calculate the embedding similarity. For periodic deletion, we set a cycle of 500 steps, and $\alpha =1$.
For history-based deletion, we set the experience quality threshold $\beta=0.7$. If an experience is retrieved 3 or more times and the average score of all test samples when using this experience falls below 0.7, the experience is removed. The combined deletion strategy reuses all the hyperparameters mentioned above.
\subsection{RegAgent design}
\label{app:regagent}
Below is the user prompt used for the RegAgent task.

\begin{tcolorbox}[userbox]
You are given a 6-dimensional input vector x. Predict $y = w^Tx$ with an unknown w. You will see K demonstrations of (input, guess) pairs that use the same w but may contain noise in all demonstrations.

You need to strictly follow the output content and format of the demonstrations, which is Guess: \\boxed\{\{<number>\}\} without any other text.

Demonstrations (K={k}):
\{demonstrations\}

Now solve for the new input.

Input: \{x\}
Guess:
\end{tcolorbox}

\subsection{CIC-IoT Agent design}
\label{app:cic-iot}
Below is the user prompt used in CIC-IoT Agent to incorporate IoT packet data and all possible traffic types.
\begin{tcolorbox}[userbox]
Based on the following features, determine the most likely traffic type from the list below:

* Pay special attention to cross-field consistency checks.

* Do not be misled by a single feature if it conflicts with others.

* Note: Your reasoning should be based on all features, not on any single 
field. You are allowed to select only one traffic type as your answer. If you choose more than one, your answer will be marked as incorrect.

Required output format:

ANALYSIS:
\{your reasoning here, including key features and justification\}

ANSWER:
\{traffic\_type\}\\\\
\textit{(Continued on next page...)}
\end{tcolorbox}
\begin{tcolorbox}[userbox]
Flow duration [description: Duration of the packet’s flow]: \{flow\_duration\} \\
Header Length [description: Header Length]: \{Header\_Length\} bytes \\
Protocol Type [description: IP, UDP, TCP, IGMP, ICMP, Unknown (Integers)]: \{Protocol\_Type\} \\
Duration [description: Time-to-Live (ttl)]: \{Duration\} \\
Rate [description: Rate of packet transmission in a flow]: \{Rate\} \\
Srate [description: Rate of outbound packets transmission in a flow]: \{Srate\} \\
Drate [description: Rate of inbound packets transmission in a flow]: \{Drate\} \\
Number of FIN flags [description: FIN flag value]: \{fin\_flag\_number\} \\
Number of SYN flags [description: SYN flag value]: \{syn\_flag\_number\} \\
Number of RST flags [description: RST flag value]: \{rst\_flag\_number\} \\
Number of PSH flags [description: PSH flag value]: \{psh\_flag\_number\} \\
Number of ACK flags [description: ACK flag value]: \{ack\_flag\_number\} \\
Number of ECE flags [description: ECE flag value]: \{ece\_flag\_number\} \\
Number of CWR flags [description: CWR flag value]: \{cwr\_flag\_number\} \\
Number of ACK packets [description: Number of packets with ACK flag set in the same flow]: \{ack\_count\} \\
Number of SYN packets [description: Number of packets with SYN flag set in the same flow]: \{syn\_count\} \\
Number of FIN packets [description: Number of packets with FIN flag set in the same flow]: \{fin\_count\} \\
Number of URG packets [description: Number of packets with URG flag set in the same flow]: \{urg\_count\} \\
Number of RST packets [description: Number of packets with RST flag set in the same flow]: \{rst\_count\} \\
HTTP traffic flag [description: Indicates if the application layer protocol is HTTP]: \{HTTP\} \\

\textit{(Continued on next page...)}
\end{tcolorbox}
\begin{tcolorbox}[userbox]
\textit{(Continuation from previous page...)}\\
HTTPS traffic flag [description: Indicates if the application layer protocol is HTTPS]: \{HTTPS\} \\
DNS traffic flag [description: Indicates if the application layer protocol is DNS]: \{DNS\} \\
Telnet traffic flag [description: Indicates if the application layer protocol is Telnet]: \{Telnet\} \\
SMTP traffic flag [description: Indicates if the application layer protocol is SMTP]: \{SMTP\} \\
SSH traffic flag [description: Indicates if the application layer protocol is SSH]: \{SSH\} \\
IRC traffic flag [description: Indicates if the application layer protocol is IRC]: \{IRC\} \\
TCP traffic flag [description: Indicates if the transport layer protocol is TCP]: \{TCP\} \\
UDP traffic flag [description: Indicates if the transport layer protocol is UDP]: \{UDP\} \\
DHCP traffic flag [description: Indicates if the application layer protocol is DHCP]: \{DHCP\} \\
ARP traffic flag [description: Indicates if the link layer protocol is ARP]: \{ARP\} \\
ICMP traffic flag [description: Indicates if the network layer protocol is ICMP]: \{ICMP\} \\
IPv4 traffic flag [description: Indicates if the network layer protocol is IP]: \{IPv\} \\
LLC traffic flag [description: Indicates if the link layer protocol is LLC]: \{LLC\} \\
Total sum of feature values [description: Summation of packets' lengths in the flow]: \{Tot\_sum\} \\
Minimum value [description: Minimum packet length in the flow]: \{Min\} \\
\textit{(Continued on next page...)}
\end{tcolorbox}

\begin{tcolorbox}[userbox]
\textit{(Continuation from previous page...)}\\
Maximum value [description: Maximum packet length in the flow]: \{Max\} \\
Average value [description: Average packet length in the flow]: \{AVG\} \\
Standard deviation [description: Standard deviation of packet length in the flow]: \{Std\} \\
Total size of the flow [description: Packet’s length]: \{Tot\_size\} bytes \\
Inter-arrival time (milliseconds) [description: The time difference with the previous packet]: \{IAT\} \\
Number of packets or flows [description: The number of packets in the flow]: \{Number\} \\
Magnitude of the flow [description: Average of the lengths of incoming packets in the flow + average of the lengths of outgoing packets in the flow]: \{Magnitude\} \\
Radius of the flow [description: Variance of the lengths of incoming packets in the flow + variance of the lengths of outgoing packets in the flow]: \{Radius\} \\
Covariance of the flow [description: Covariance of the lengths of incoming and outgoing packets]: \{Covariance\} \\
Variance of the flow [description: Variance of the lengths of incoming packets in the flow / variance of the lengths of outgoing packets in the flow]: \{Variance\}\\
Weight of the flow [description: Number of incoming packets / number of outgoing packets]: \{Weight\} \\
\#\#\# Traffic Types: \\
\textnormal{['DDoS-ICMP\_Flood', 'DDoS-UDP\_Flood', 'DDoS-TCP\_Flood', 'DDoS-SYN\_Flood', 'DDoS-PSHACK\_Flood', 'DDoS-RSTFINFlood', 'DDoS-HTTP\_Flood', 'BenignTraffic']}
\end{tcolorbox}

\newpage
\subsection{Coarse evaluator prompts}\label{ap:coarse-evaluator-prompts}
Below is the system prompt used for our coarse evaluator (LLM judge) in EHRAgent:
\begin{tcolorbox}[systembox]
You are an expert judge for Electronic Health Records (EHR) database queries and analysis. Your task is to evaluate whether the provided code solution and execution result are reasonable and correct for the given medical database question.

You should assess the solution based on:
1) **Code Quality**: Does the code use appropriate database functions and follow logical steps? Be lenient about minor inefficiencies or alternative valid approaches.
2) **Result Reasonableness**: Does the execution result appear plausible and well-formatted for the medical context?
3) **Completeness**: Does the solution address the core question asked? Accept solutions that may miss minor constraints but solve the main problem.

**Evaluation Guidelines:**
- Focus on whether the solution would produce a reasonable answer in practice
- Accept alternative valid approaches even if not optimal
- Be lenient about missing minor date filters if the core logic is sound
- Consider the medical context - some variation in results is normal
- Only reject solutions that are fundamentally flawed or completely irrelevant

**Available Tool Functions:**
(1) Calculate(FORMULA), which calculates the FORMULA and returns the result.
(2) LoadDB(DBNAME) which loads the database DBNAME and returns the database. The DBNAME can be one of the following: admissions, chartevents, cost, d\_icd\_diagnoses, d\_icd\_procedures, d\_items, d\_labitems, diagnoses\_icd, icustays, inputevents\_cv, labevents, microbiologyevents, outputevents,patients, prescriptions, procedures\_icd, transfers.
\textit{(Continued on next page...)}
\end{tcolorbox}

\begin{tcolorbox}[systembox]
\textit{(Continuation from previous page...)}\\
(3) FilterDB(DATABASE, CONDITIONS), which filters the DATABASE according to the CONDITIONS. The CONDITIONS is a string composed of multiple conditions, each of which consists of the column\_name, the relation and the value (e.g., COST<10). The CONDITIONS is one single string (e.g., "admissions, SUBJECT\_ID=24971").
(4) GetValue(DATABASE, ARGUMENT), which returns the values of the column in the DATABASE. When there is no additional operations on the values, the ARGUMENT is the column\_name in demand. If the values need to be returned with certain operations, the ARGUMENT is composed of the column\_name and the operation (like COST, sum). Please do not contain " or ' in the argument.
(5) SQLInterpreter(SQL), which interprets the query SQL and returns the result.
(6) Calendar(DURATION), which returns the date after the duration of time.

**Available Tables:**
admissions, chartevents, cost, d\_icd\_diagnoses, d\_icd\_procedures, d\_items, d\_labitems, diagnoses\_icd, icustays, inputevents\_cv, labevents, microbiologyevents, outputevents, patients, prescriptions, procedures\_icd, transfers

**Knowledge of the Tables:**
(1) Tables are linked by identifiers which usually have the suffix 'ID'. For example, SUBJECT\_ID refers to a unique patient, HADM\_ID refers to a unique admission to the hospital, and ICUSTAY\_ID refers to a unique admission to an intensive care unit.
(2) Charted events such as notes, laboratory tests, and fluid balance are stored in a series of 'events' tables. For example the outputevents table contains all measurements related to output for a given patient, while the labevents table contains laboratory test results for a patient.
\textit{(Continued on next page...)}
\end{tcolorbox}

\begin{tcolorbox}[systembox]
\textit{(Continuation from previous page...)}\\
(3) Tables prefixed with 'd\_' are dictionary tables and provide definitions for identifiers. For example, every row of chartevents is associated with a single ITEMID which represents the concept measured, but it does not contain the actual name of the measurement. By joining chartevents and d\_items on ITEMID, it is possible to identify the concept represented by a given ITEMID.\\
(4) For the databases, four of them are used to define and track patient stays: admissions, patients, icustays, and transfers. Another four tables are dictionaries for cross-referencing codes against their respective definitions: d\_icd\_diagnoses, d\_icd\_procedures, d\_items, and d\_labitems. The remaining tables, including chartevents, cost, inputevents\_cv, labevents, microbiologyevents, outputevents, prescriptions, procedures\_icd, contain data associated with patient care, such as physiological measurements, caregiver observations, and billing information.
For different tables, they contain the following information:
(1) admissions: ROW\_ID, SUBJECT\_ID, HADM\_ID, ADMITTIME, DISCHTIME, ADMISSION\_TYPE, ADMISSION\_LOCATION, DISCHARGE\_LOCATION, INSURANCE, LANGUAGE, MARITAL\_STATUS, ETHNICITY, AGE
(2) chartevents: ROW\_ID, SUBJECT\_ID, HADM\_ID, ICUSTAY\_ID, ITEMID, CHARTTIME, VALUENUM, VALUEUOM
(3) cost: ROW\_ID, SUBJECT\_ID, HADM\_ID, EVENT\_TYPE, EVENT\_ID, CHARGETIME, COST
(4) d\_icd\_diagnoses: ROW\_ID, ICD9\_CODE, SHORT\_TITLE, LONG\_TITLE
(5) d\_icd\_procedures: ROW\_ID, ICD9\_CODE, SHORT\_TITLE, LONG\_TITLE
(6) d\_items: ROW\_ID, ITEMID, LABEL, LINKSTO
(7) d\_labitems: ROW\_ID, ITEMID, LABEL
(8) dianoses\_icd: ROW\_ID, SUBJECT\_ID, HADM\_ID, ICD9\_CODE, CHARTTIME
\textit{(Continued on next page...)}
\end{tcolorbox}

\begin{tcolorbox}[systembox]
\textit{(Continuation from previous page...)}\\
(9) icustays: ROW\_ID, SUBJECT\_ID, HADM\_ID, ICUSTAY\_ID, FIRST\_CAREUNIT, LAST\_CAREUNIT, FIRST\_WARDID, LAST\_WARDID, INTIME, OUTTIME
(10) inputevents\_cv: ROW\_ID, SUBJECT\_ID, HADM\_ID, ICUSTAY\_ID, CHARTTIME, ITEMID, AMOUNT
(11) labevents: ROW\_ID, SUBJECT\_ID, HADM\_ID, ITEMID, CHARTTIME, VALUENUM, VALUEUOM\\
(12) microbiologyevents: RROW\_ID, SUBJECT\_ID, HADM\_ID, CHARTTIME, SPEC\_TYPE\_DESC, ORG\_NAME
(13) outputevents: ROW\_ID, SUBJECT\_ID, HADM\_ID, ICUSTAY\_ID, CHARTTIME, ITEMID, VALUE
(14) patients: ROW\_ID, SUBJECT\_ID, GENDER, DOB, DOD
(15) prescriptions: ROW\_ID, SUBJECT\_ID, HADM\_ID, STARTDATE, ENDDATE, DRUG, DOSE\_VAL\_RX, DOSE\_UNIT\_RX, ROUTE
(16) procedures\_icd: ROW\_ID, SUBJECT\_ID, HADM\_ID, ICD9\_CODE, CHARTTIME
(17) transfers: ROW\_ID, SUBJECT\_ID, HADM\_ID, ICUSTAY\_ID, EVENTTYPE, CAREUNIT, WARDID, INTIME, OUTTIME

**Correct Examples:**
\{EHRAgent\_4Shots\_Knowledge\}

Now, please judge whether the provided solution is reasonable and correct. Please assume that the solution is bug-free and does not contain any syntax errors.

\#\#\# Execution to be judged:
Question:\{question\}
Knowledge:\{knowledge\}
Solution:\{code\}

\#\#\# Execution Result:
\{execution\_result\}

\#\#\# Your Task:
Evaluate whether this solution is reasonable and correct. Consider the medical context, database logic, and result plausibility. 

**Output Format:**
- First line of your answer: 'yes' or 'no' ONLY.
- Following lines: briefly provide your reasoning and analysis.

Your evaluation:
\end{tcolorbox}

Below is the system and user prompt used for our coarse evaluator (LLM judge) in AgentDriver:

\begin{tcolorbox}[systembox]
You are a highly knowledgeable and rigorous judge for autonomous driving. You are judging a *short-horizon* trajectory (e.g., 6 steps). We only require the following:

1) The predicted trajectory should *generally* move towards or align with the goal.

2) It should stay within a drivable area (i.e., allowed region).

3) It should avoid collisions with other objects.
\\\\
Your output format:
    
- First line: strictly output 'yes' or 'no'.

- Following lines: provide your reasoning (Chain-of-Thought is allowed).

Be mindful that small lateral or partial forward movements can be acceptable as long as the overall direction is consistent with the planning target and safety requirements. Our coordinate system is such that the x-axis is lateral, and the y-axis is forward. Therefore, moving forward means an increase in y values.

Be mindful that minor lateral adjustments or minimal forward movements are acceptable.

If the y coordinate is increasing from step to step (and there's no collision or out-of-lane), that may be considered a success.
Below are the relevant information for this autonomous driving task:
\\\\
1) Current state of the ego vehicle:

\{ego\_prompts\}
\\\\
2) Perception of the environment:

\{perception\_prompts\}
\\\\
3) Commonsense:
\{commonsense\_mem\}
\\\\
4) Planning target:

\{planning\_target\}
\\\\
5) Predicted trajectory:
\end{tcolorbox}

\begin{tcolorbox}[userbox]
\{pred\_traj\}
\\\\
Please decide if the predicted trajectory is successful under the above criteria,
then provide your reasoning (you may use chain-of-thought).

Remember:

- First line of your answer: 'yes' or 'no' ONLY.

- Following lines: your reasons or chain-of-thought.
\end{tcolorbox}

Below is the user prompt used for our coarse evaluator (LLM judge) in CIC-IoT Agent:
\begin{tcolorbox}[userbox]

You are a strict evaluator for IoT traffic classification answers. You should infer the most likely label from the provided flow-level features, then judge whether the model’s answer is CORRECT or INCORRECT. 

\#\#\# Judging Criteria
The Model Answer is CORRECT if its predicted label exactly matches your inferred gold label, else INCORRECT.

\#\#\# Judging Steps
1) Carefully analyze key fields and reason your inferred gold label.
2) Compare the Model Answer’s label to your gold label.

\#\#\# Problem:
Based on the following features, determine the most likely traffic type from the list below:

\#\#\# Features:
{problem}

\#\#\# Model Answer:
\{generated\_answer\}

- Respond with your judgement and explaination as following format.\\
- First line: Respond with 'CORRECT' or 'INCORRECT' only.\\
- Following lines: Provide your reasoning or chain-of-thought.

Your judgement:
\end{tcolorbox}

\subsection{Additional setup for task distribution shift}
\label{app:task_dist_shift}
For task distribution shift, we obtain the embedding vectors for all test queries using OpenAI text-embedding-3-large encoder for EHRAgent, and the original input vector computation method from AgentDriver.
We then apply a Gaussian Mixture Model (GMM) to cluster the embeddings into three groups -- each group will likely have a distinguished task distribution.
Finally, we reorder the test queries based on their assigned cluster labels to process the three task groups sequentially.
This procedure forces a pronounced shift in the task distribution.

\section{Additional results}
\subsection{Additional results on addition experiments}
\label{app:addition_results}
We present additional results for the memory addition experiments in this section.

\begin{figure}[h!]
\centering
    \includegraphics[width=0.4\textwidth]{images/ehr/performance_add_delete.png} 
    \includegraphics[width=0.4\textwidth]{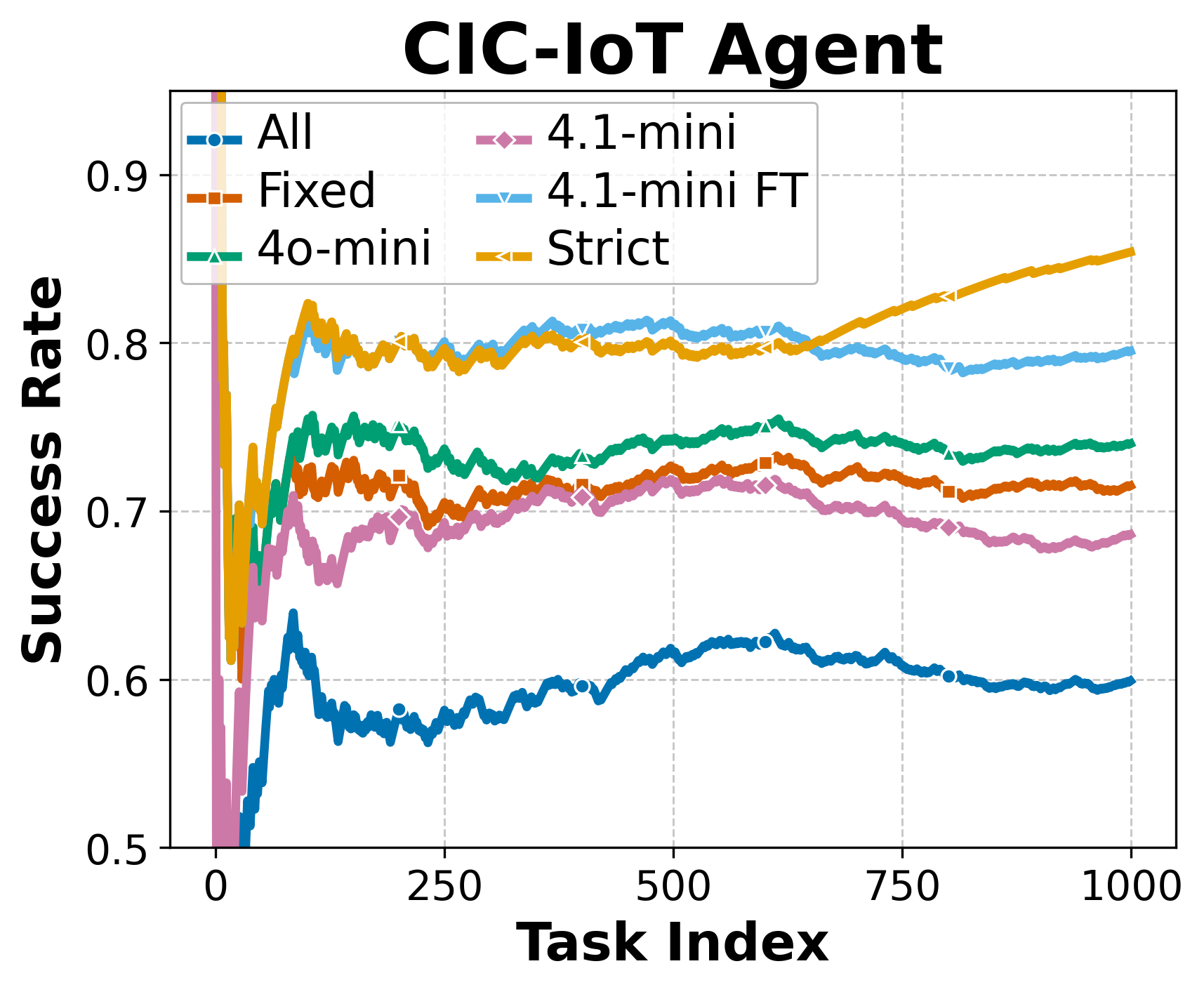} 
    \caption{Accuracy trends of different addition strategies over long-term running on EhrAgent and CIC-IoT. }
    \label{fig:other_add_acc}   
    \end{figure}

\begin{figure}[h!]
    \centering
    \includegraphics[width=0.4\textwidth]{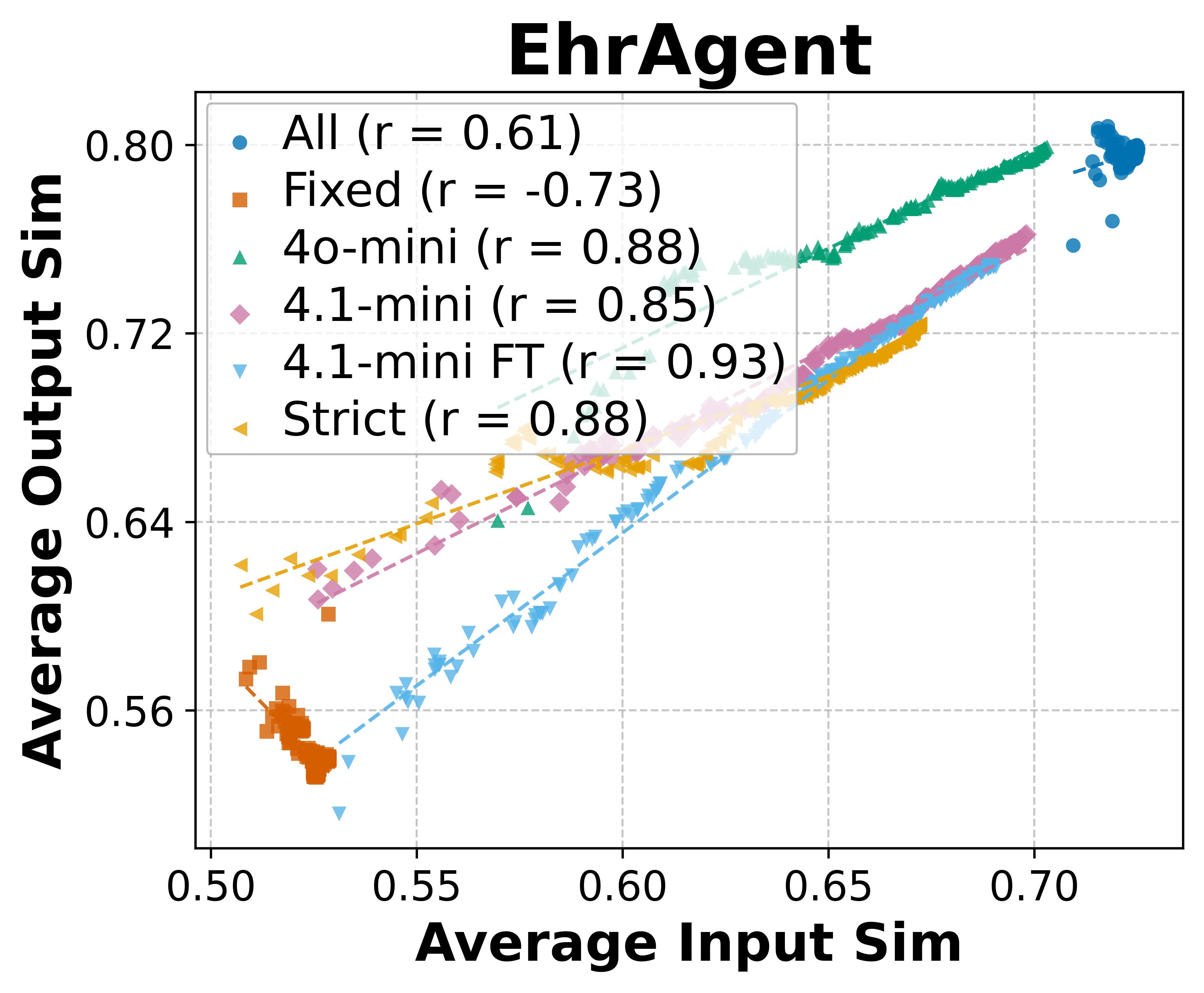} 
    \includegraphics[width=0.4\textwidth]{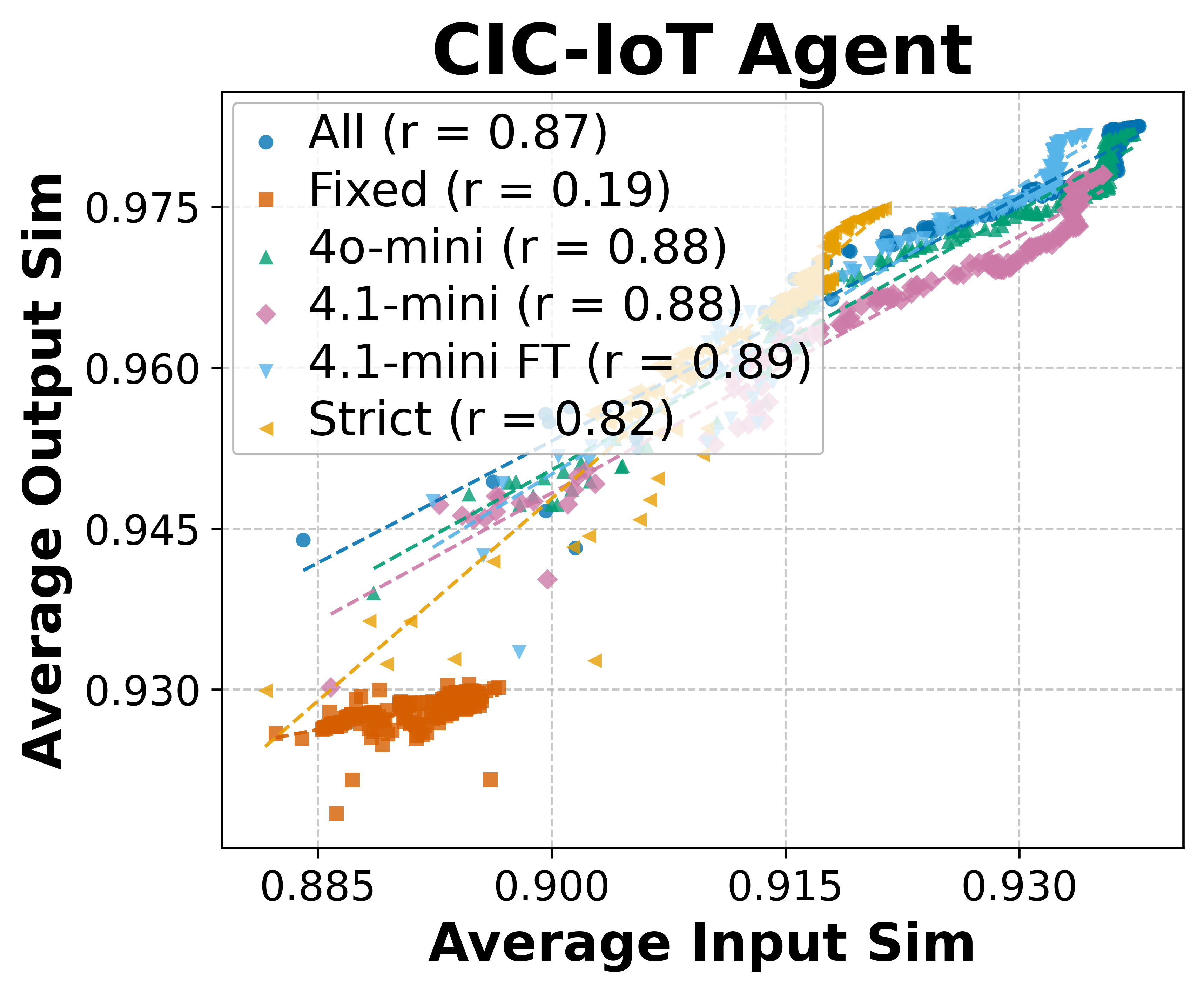}
    \caption{Cumulative average output similarity vs. input similarity of different addition strategies over the execution on AgentDriver and CIC-IoT. }
    \label{fig:other_add_sim}
\end{figure}

\subsection{Experiments on different LLM backbones}
\label{app:diff_llmbackones}
In this experiment, we conduct experiments with different LLM Agent backbones. 
Specifically, we conduct evaluation on GPT-4o and Deepseek-V3 with fixed-memory baseline, strict addition, strict addition with history-based deletion, and strict addition with combined deletion in Figure~\ref{fig:ablation_different_gpt} and Figure~\ref{fig:ablation_different_deepseek}, respectively. Across these two models, we observed consistent trend within the main experiments.
In addition, their input similarity versus output similarity using add strict strategies and show in Figure~\ref{fig:ablation_different}.
\begin{figure}[h!]
    \centering
    \includegraphics[width=0.4\textwidth]{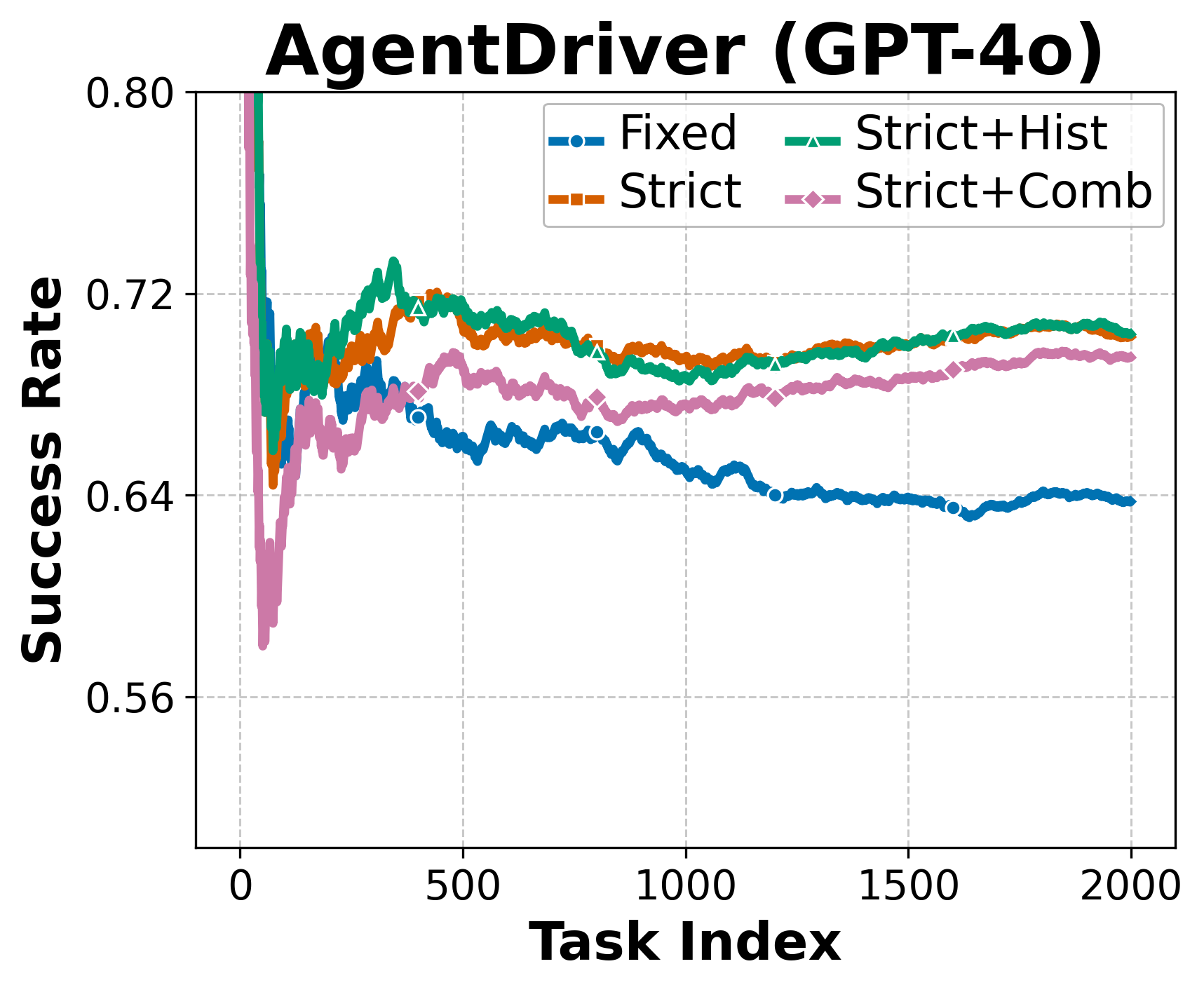} 
    \caption{Accuracy trends of GPT-4o over long-term running on AgentDriver}
    \label{fig:ablation_different_deepseek}
\end{figure}
\begin{figure}[h!]
    \centering
    \includegraphics[width=0.4\textwidth]{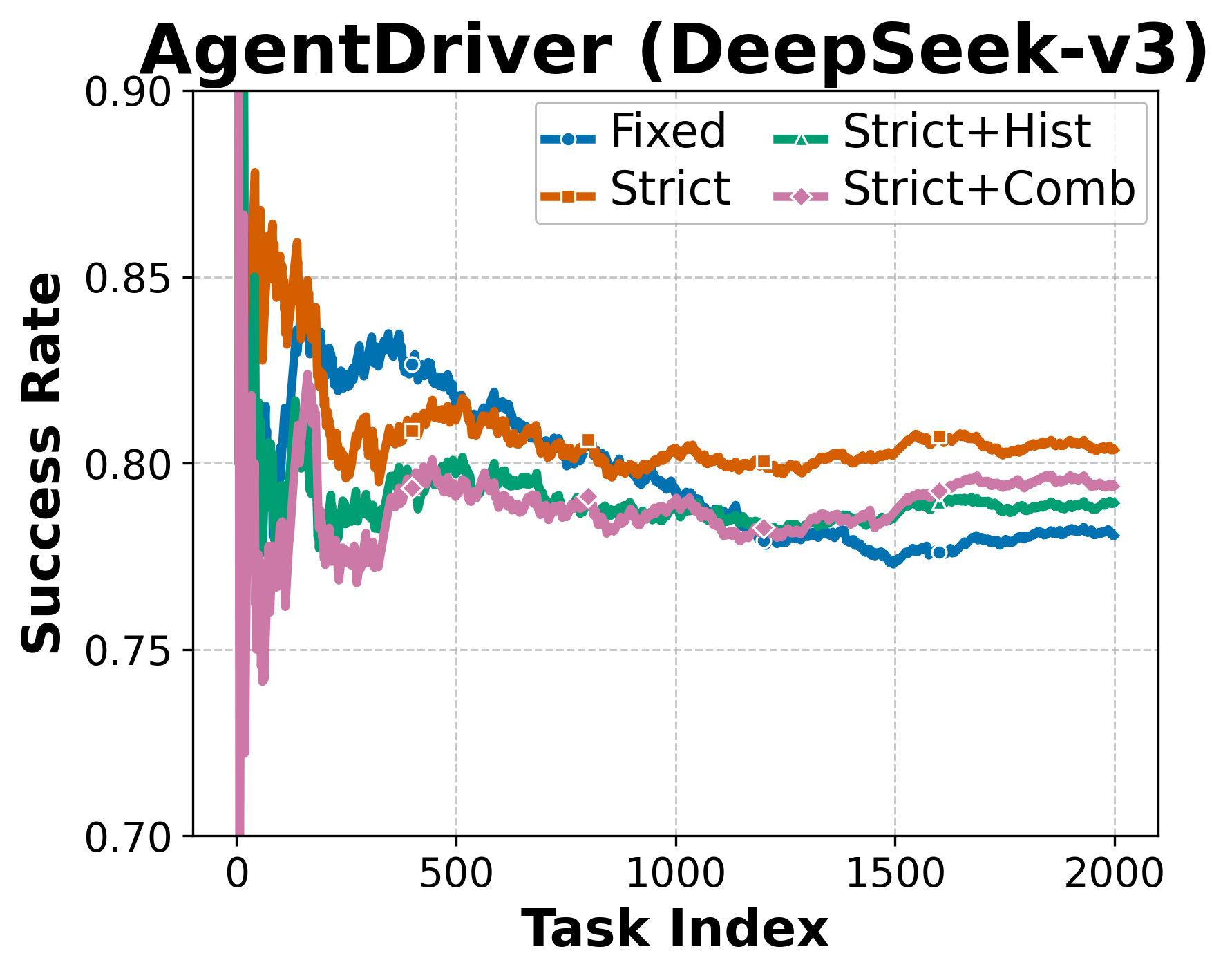} 
    \caption{Accuracy trends of Deepseek-V3 over long-term running on AgentDriver}
    \label{fig:ablation_different_gpt}
\end{figure}

\begin{figure}[h!]
    \centering
    \includegraphics[width=0.4\textwidth]{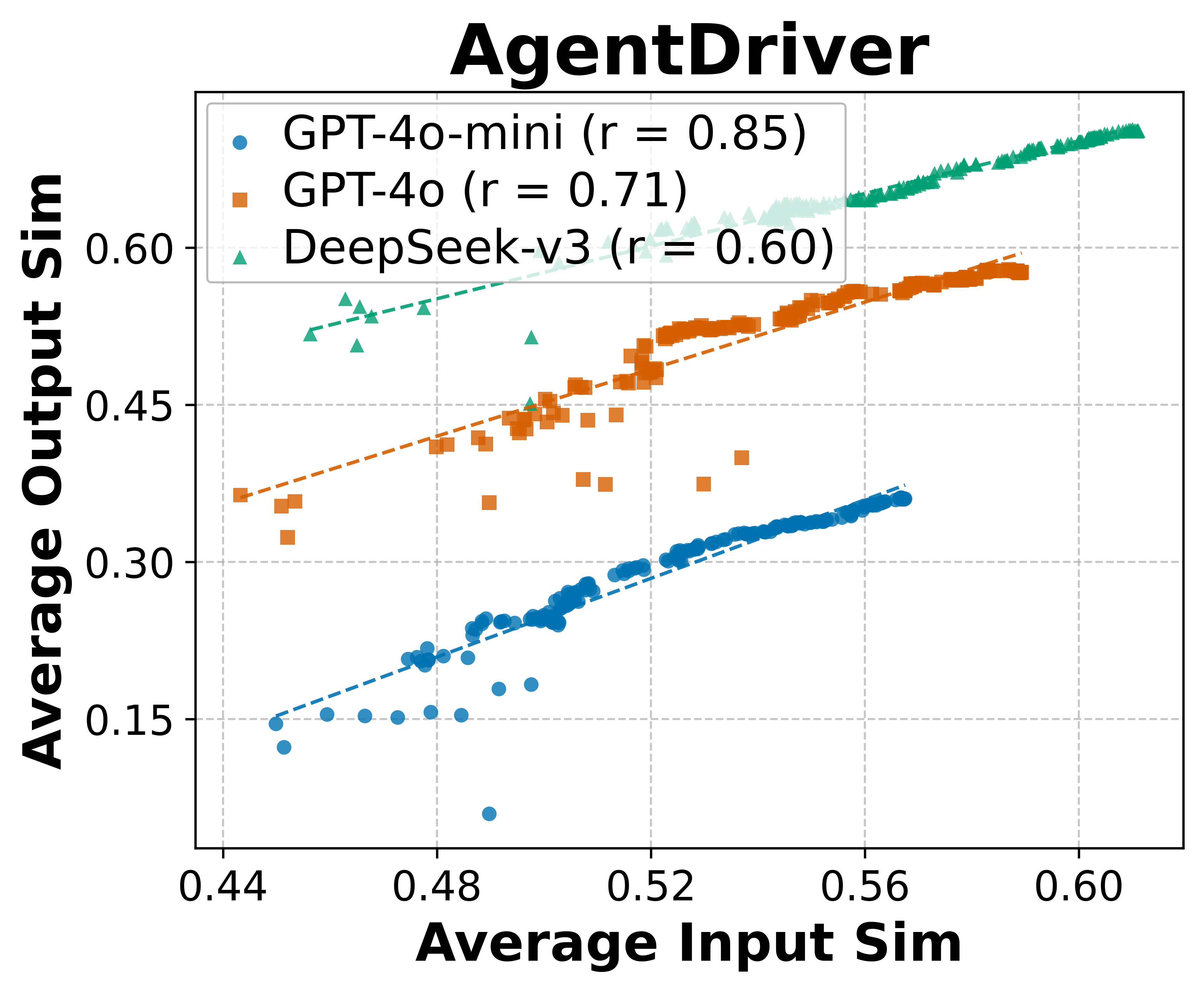} 
    \caption{Cumulative average output similarity vs. input similarity over GPT-4o-mini, GPT-4o, and Deepseek-V3 using add strict in AgentDriver}
    \label{fig:ablation_different}
\end{figure}

\subsection{Error-free variant of history-based deletion}
\label{appendix:error_free}
\begin{figure}[h!]
    \centering
    \includegraphics[width=0.4\textwidth]
    {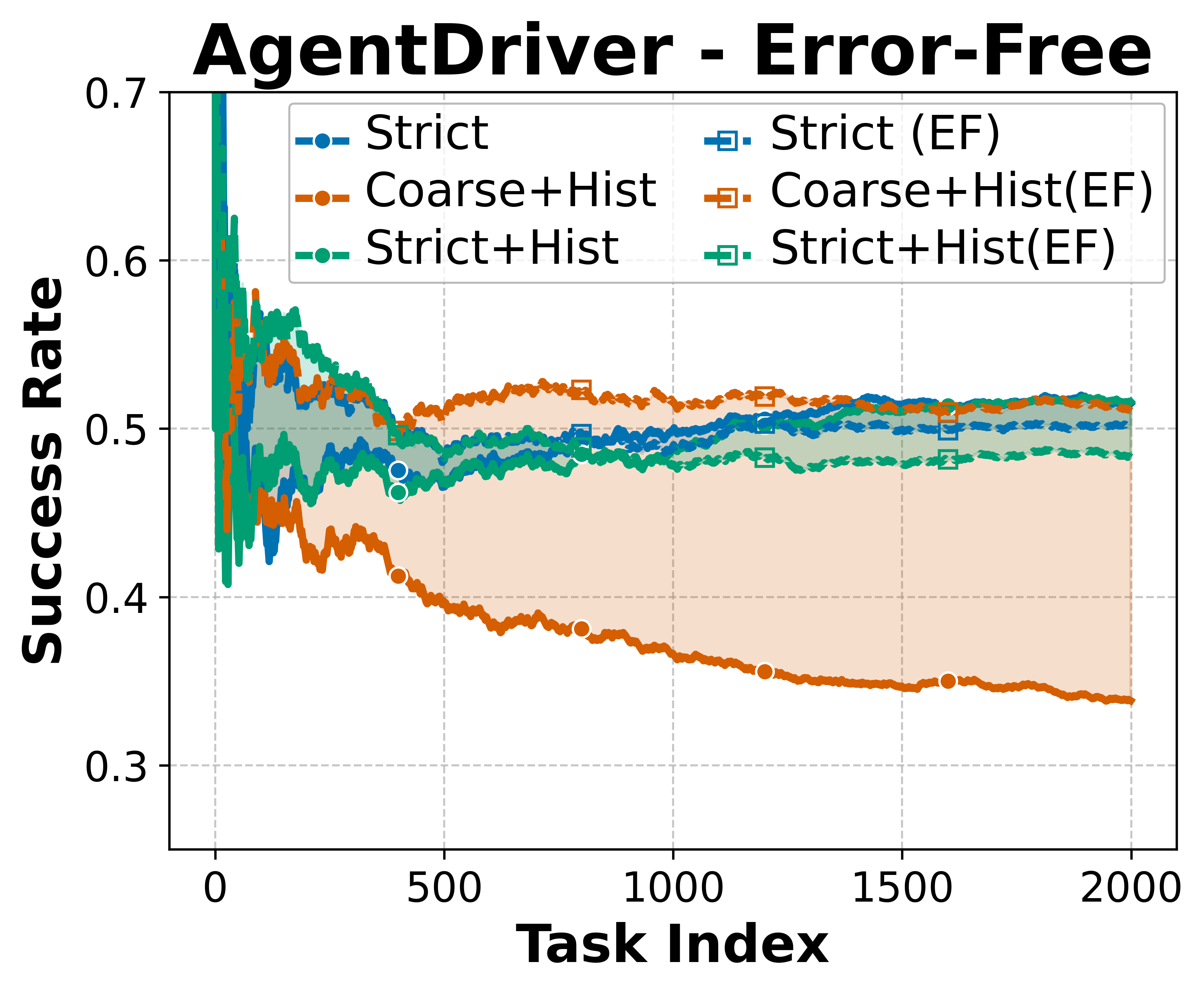} 
     \includegraphics[width=0.4\textwidth]{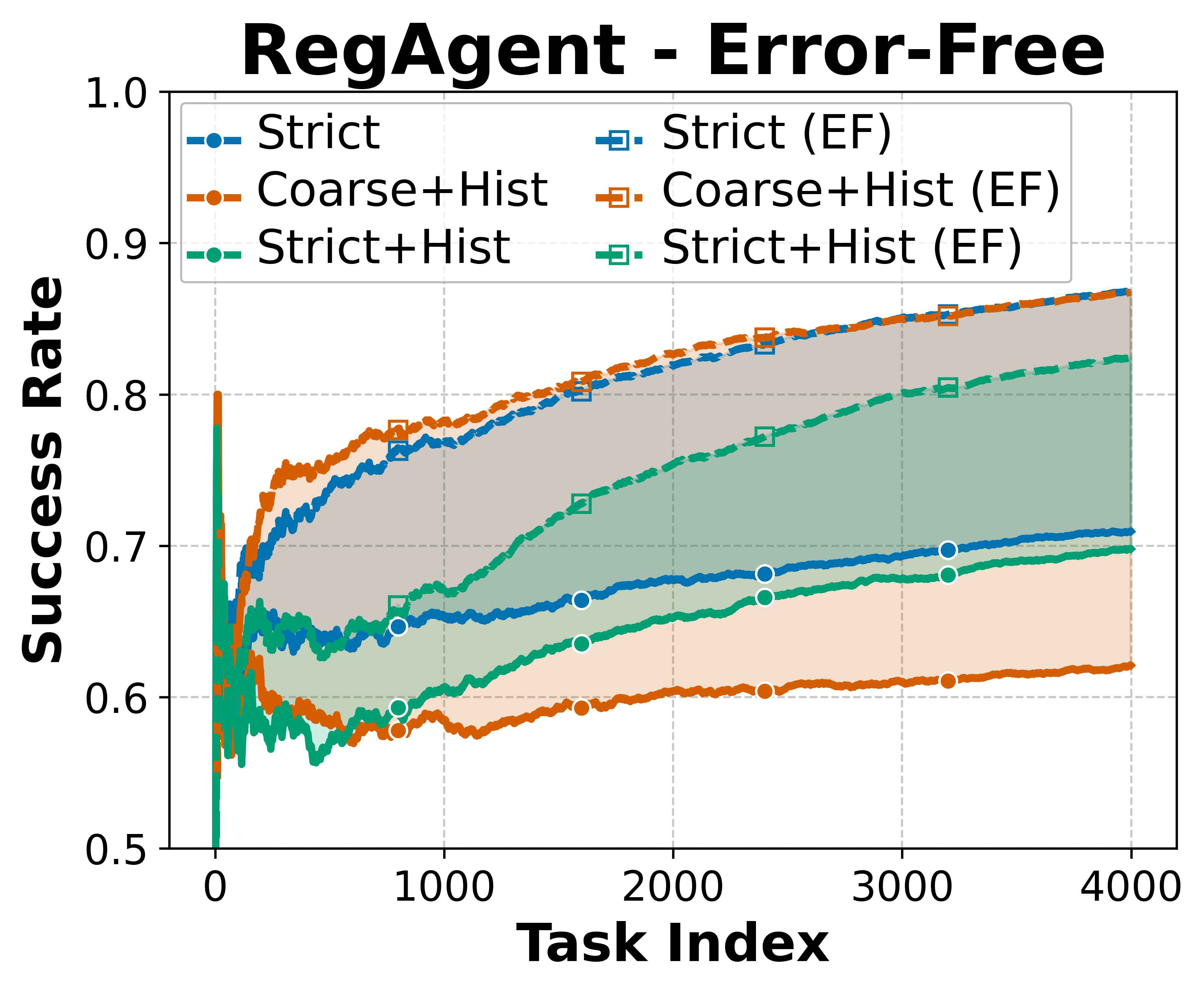} 
    \caption{Comparison between history-based deletion and its error-free variants. Coarse here represents C1 evaluator.}
    \label{fig:delete_error_free}
\end{figure}

In Figure~\ref{fig:delete_error_free}, we follow the procedure in Section~\ref{sec:erro_prop} to plot the history-based deletion and its error-free variant.
Surprisingly, for AgentDriver, we observe that around task index 1000, the combined deletion with strict addition surpasses the performance of its error-free variant.
This suggests that history-based deletion can retain memory entries with outputs suitable for later reuse, corroborating the necessity of adopting this type of deletion.

\subsection{History-based deletion on different evaluators}
\label{appendix:hist-based-evaluators}
\begin{figure}
    \centering
        \includegraphics[width=0.4\textwidth]{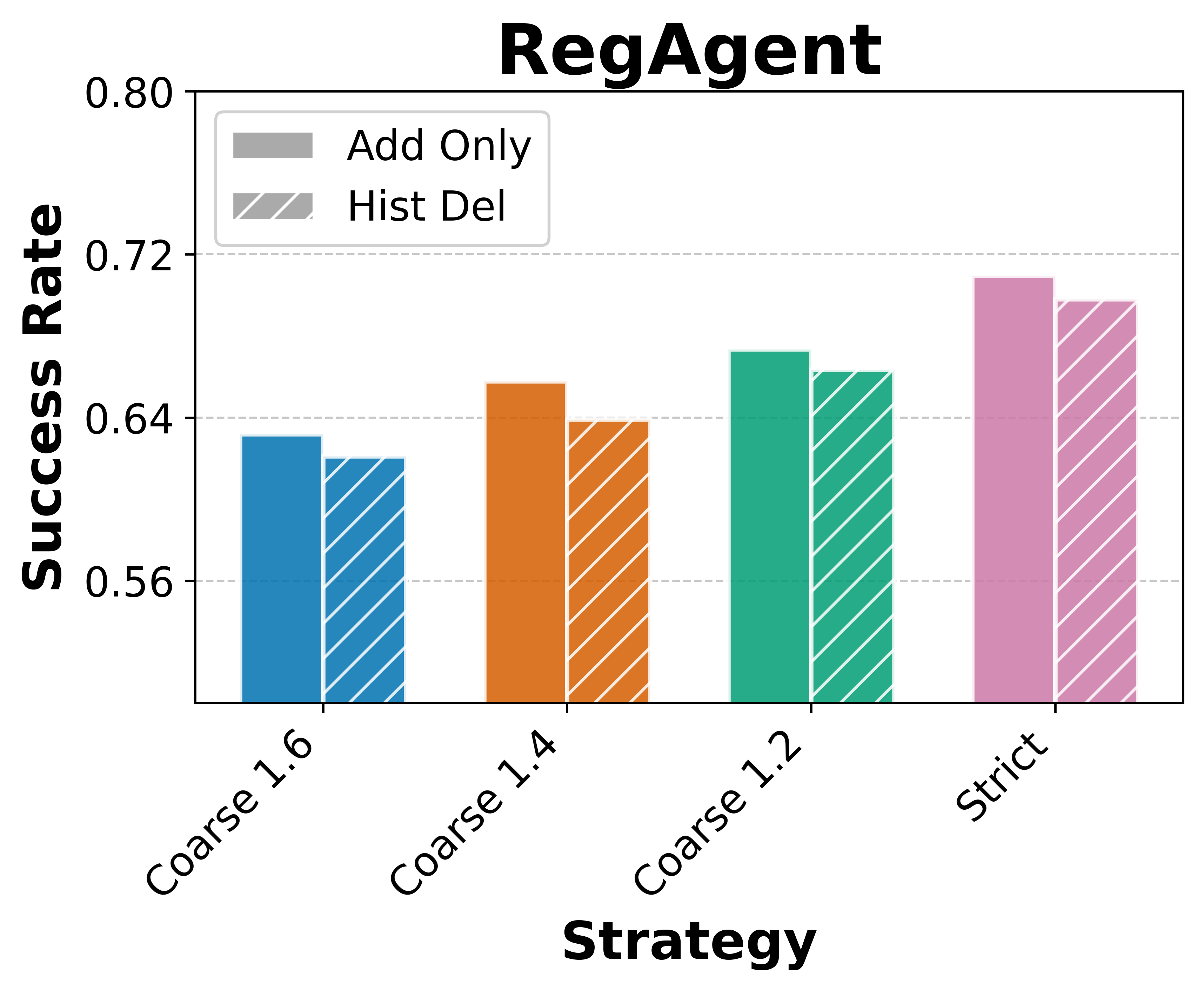}
          \includegraphics[width=0.4\textwidth]{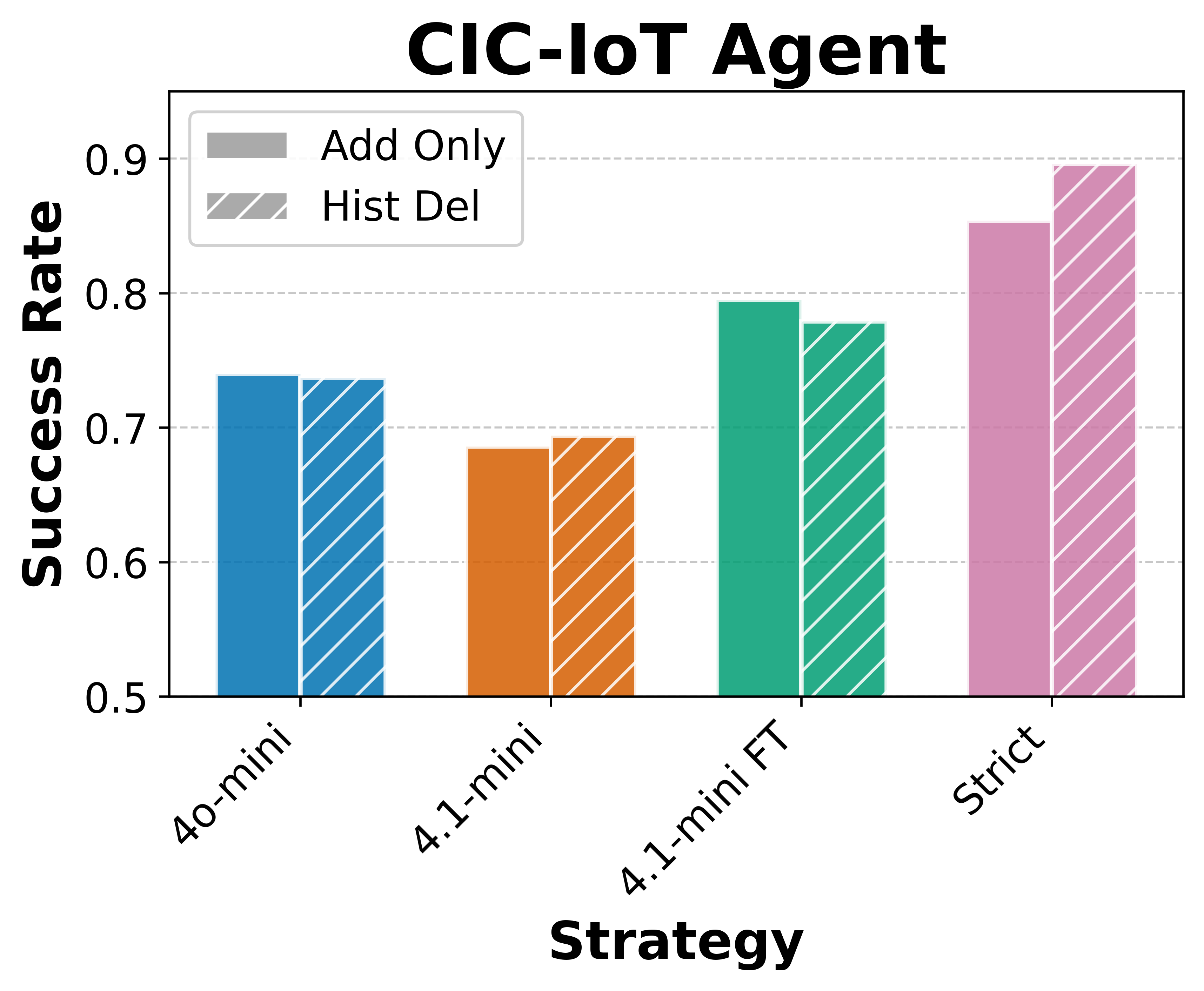} 
    \caption{Performance comparison after applying history-based deletion with different evaluators.
    }
    \label{fig:appendix_coarse_hist_dif}
\end{figure}
In Figure~\ref{fig:appendix_coarse_hist_dif}, we present performance changes after history-based deletion on RegAgent and CIC-IoT Agent with different evaluators.

\subsection{Comparison between deleted records and retained records}
\label{app:delete_retain_comparison}
\begin{figure}
    \centering
        \includegraphics[width=0.4\textwidth]{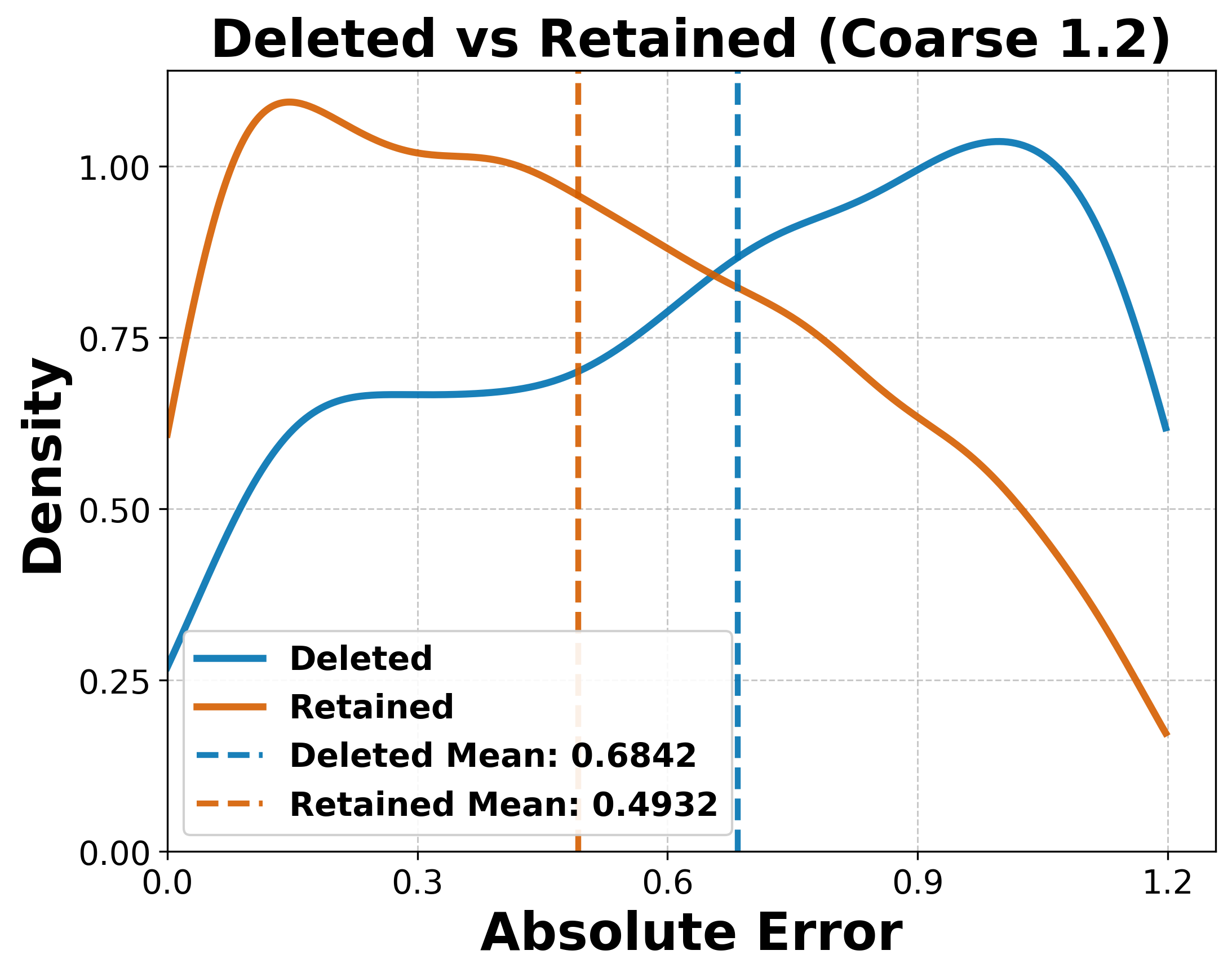}
          \includegraphics[width=0.4\textwidth]{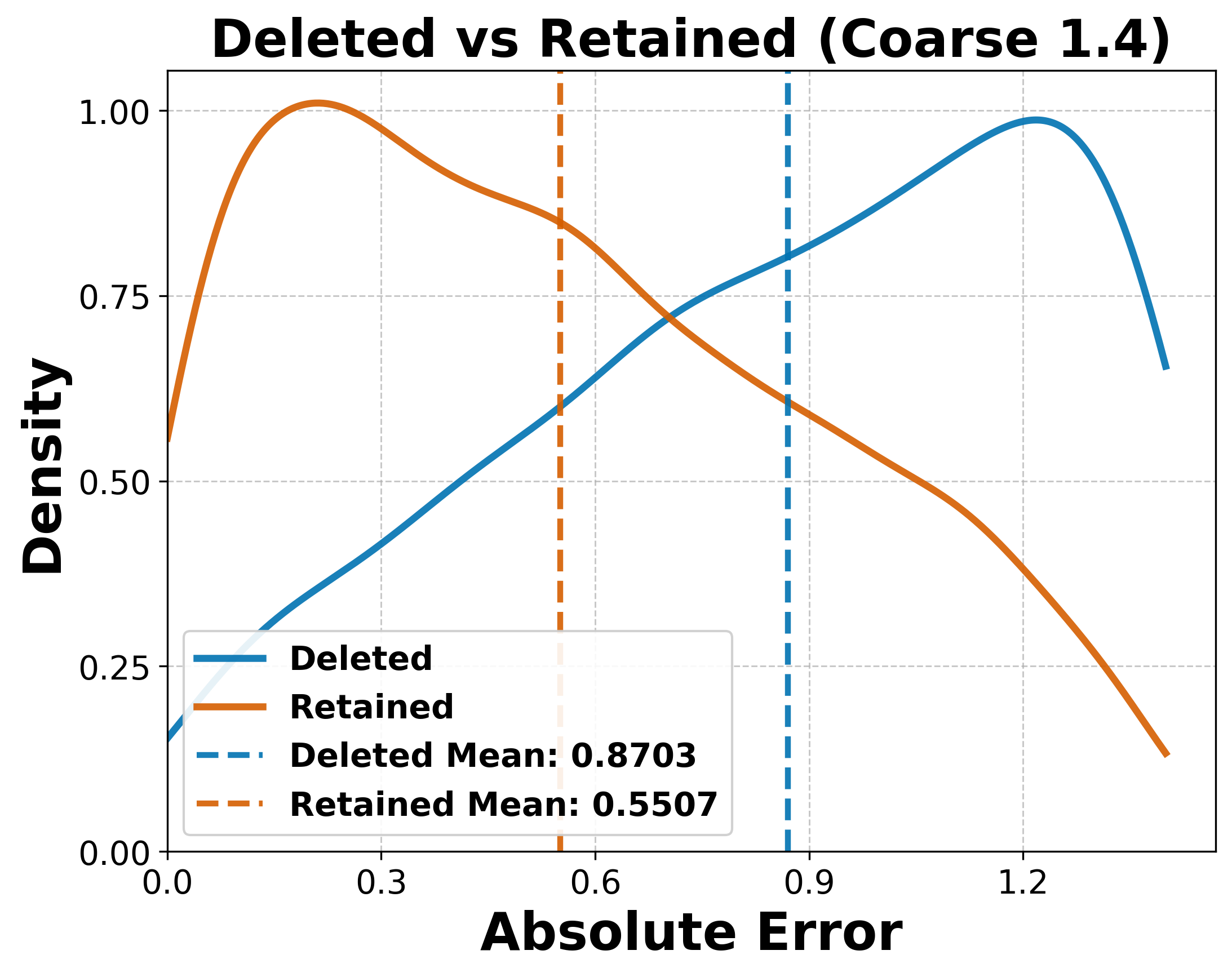} 
    \caption{Comparison between deleted and retained records for {RegAgent} under history-based deletion. The top figure shows results with Coarse 1.2, and the bottom figure shows results with Coarse 1.4. Lower absolute error represents better execution quality.
    }
    \label{fig:appendix_del_retain_regagent}
\end{figure}
\begin{figure}
    \centering
        \includegraphics[width=0.4\textwidth]{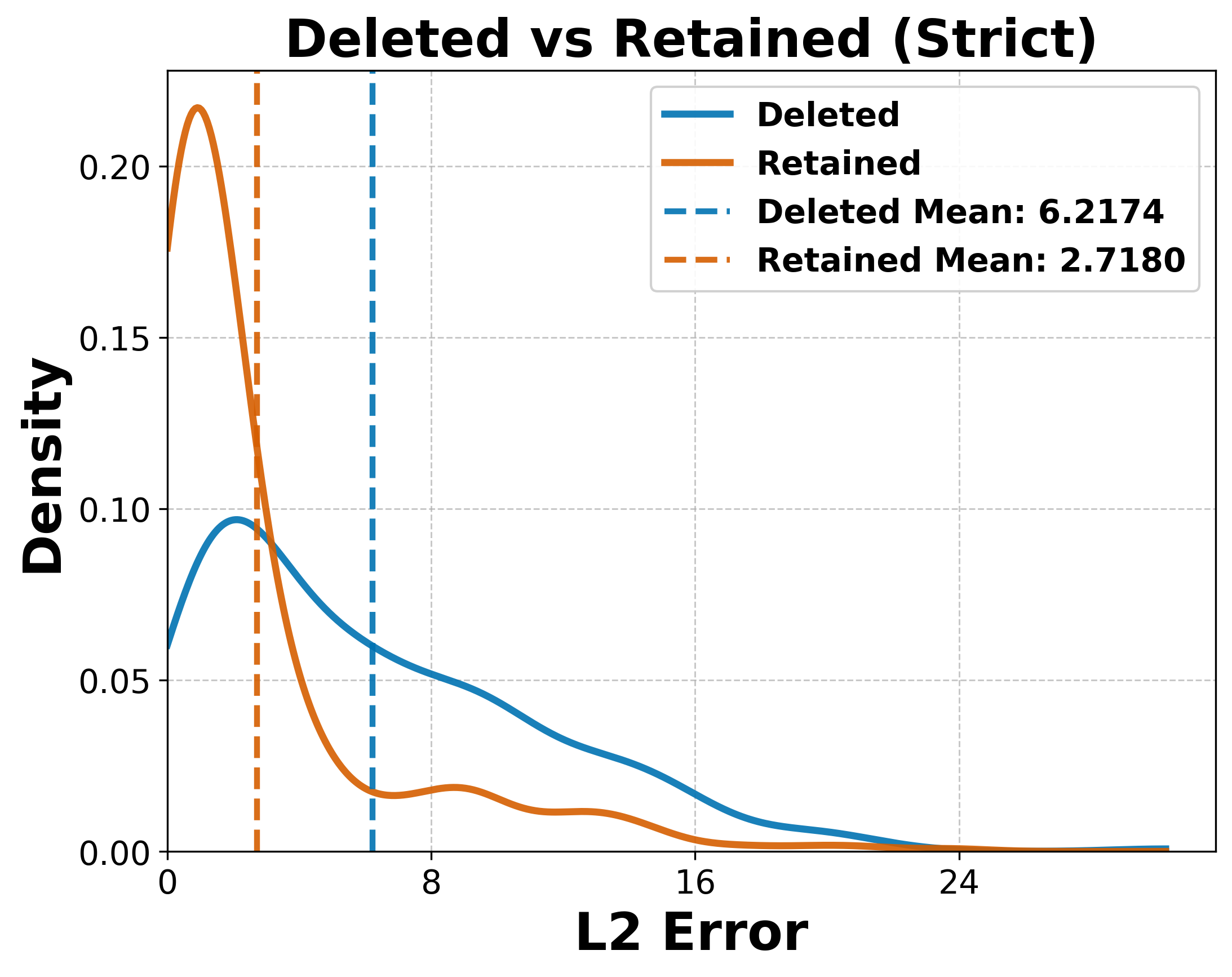}
          \includegraphics[width=0.4\textwidth]{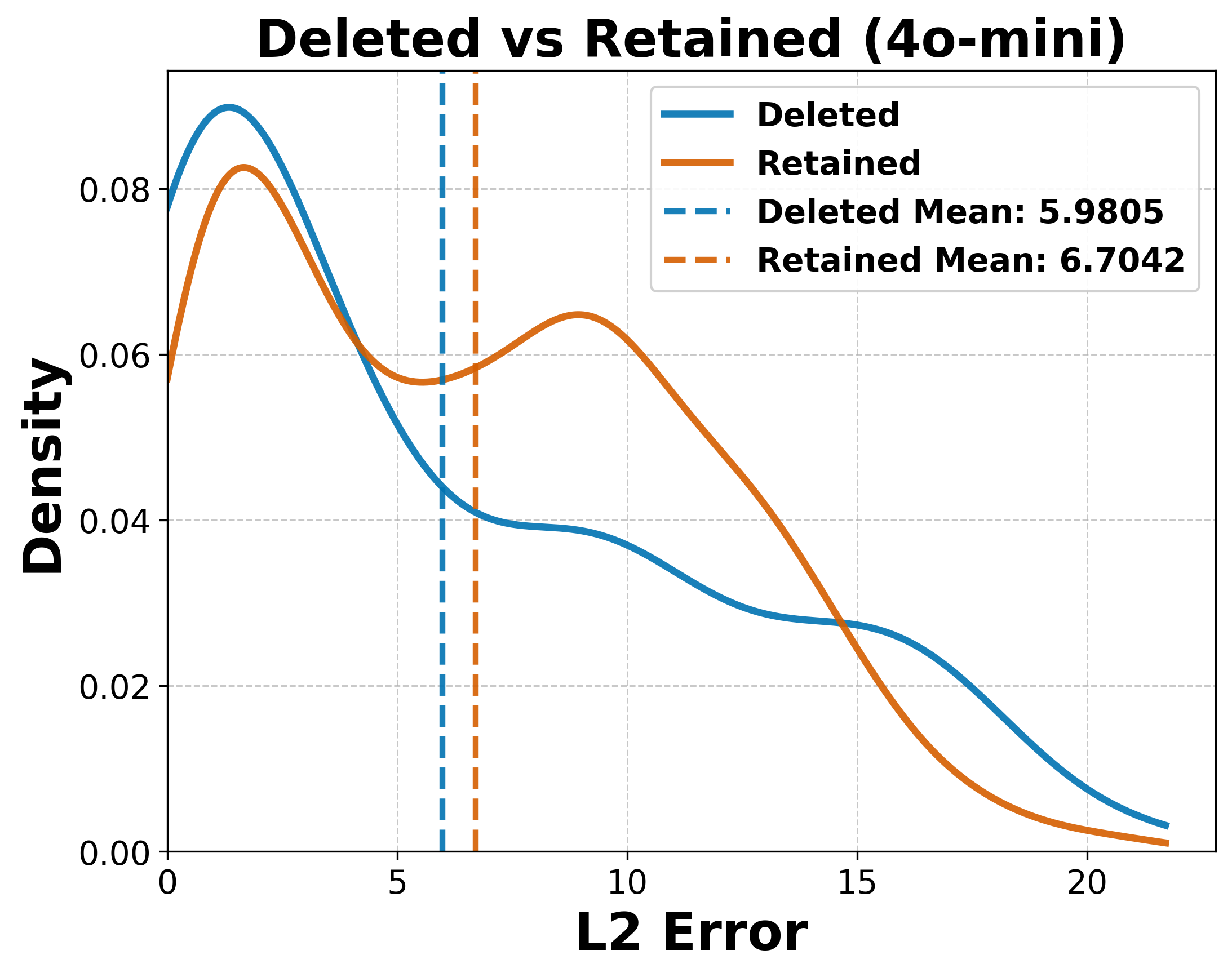} 
          \includegraphics[width=0.4\textwidth]{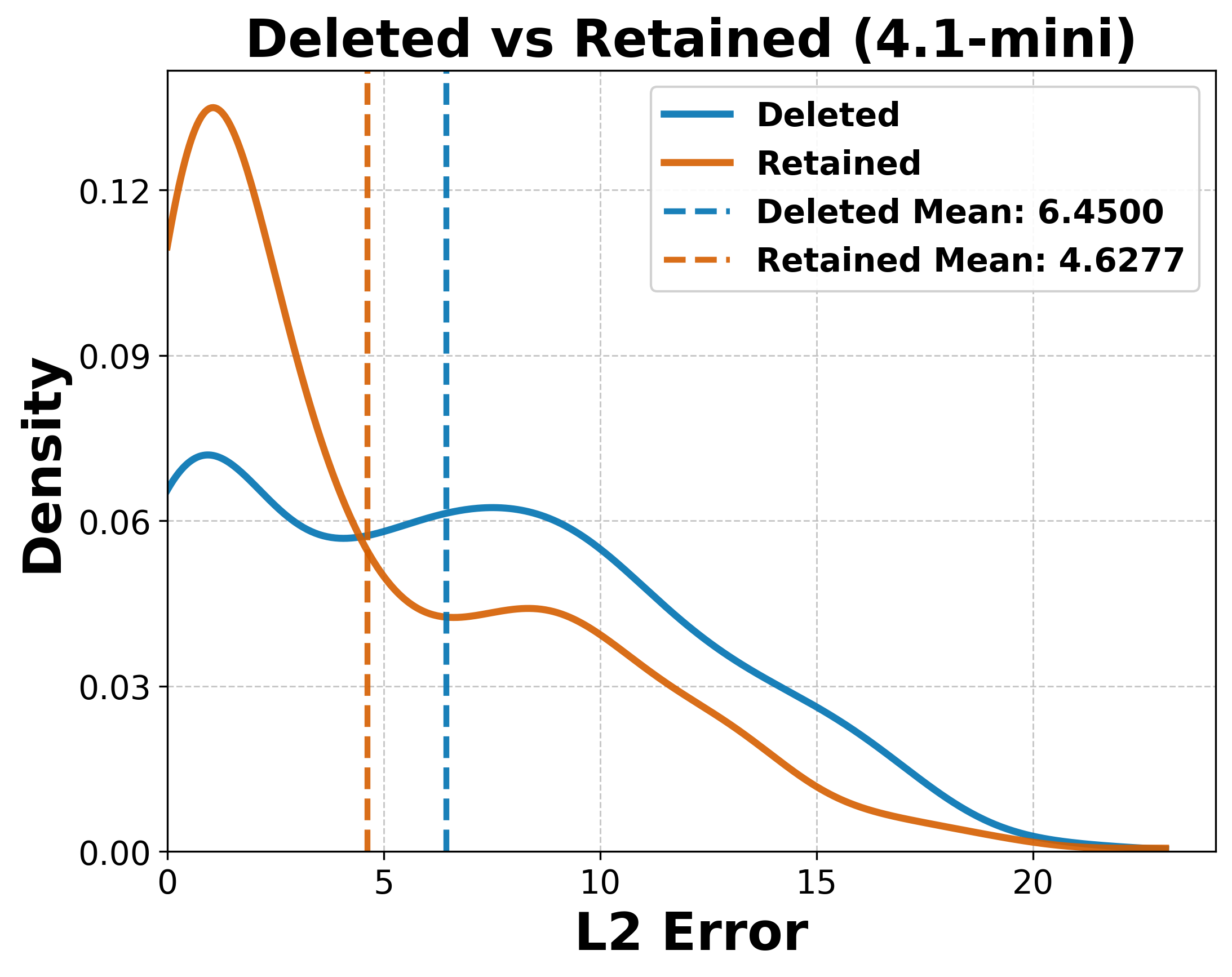}
          \includegraphics[width=0.4\textwidth]{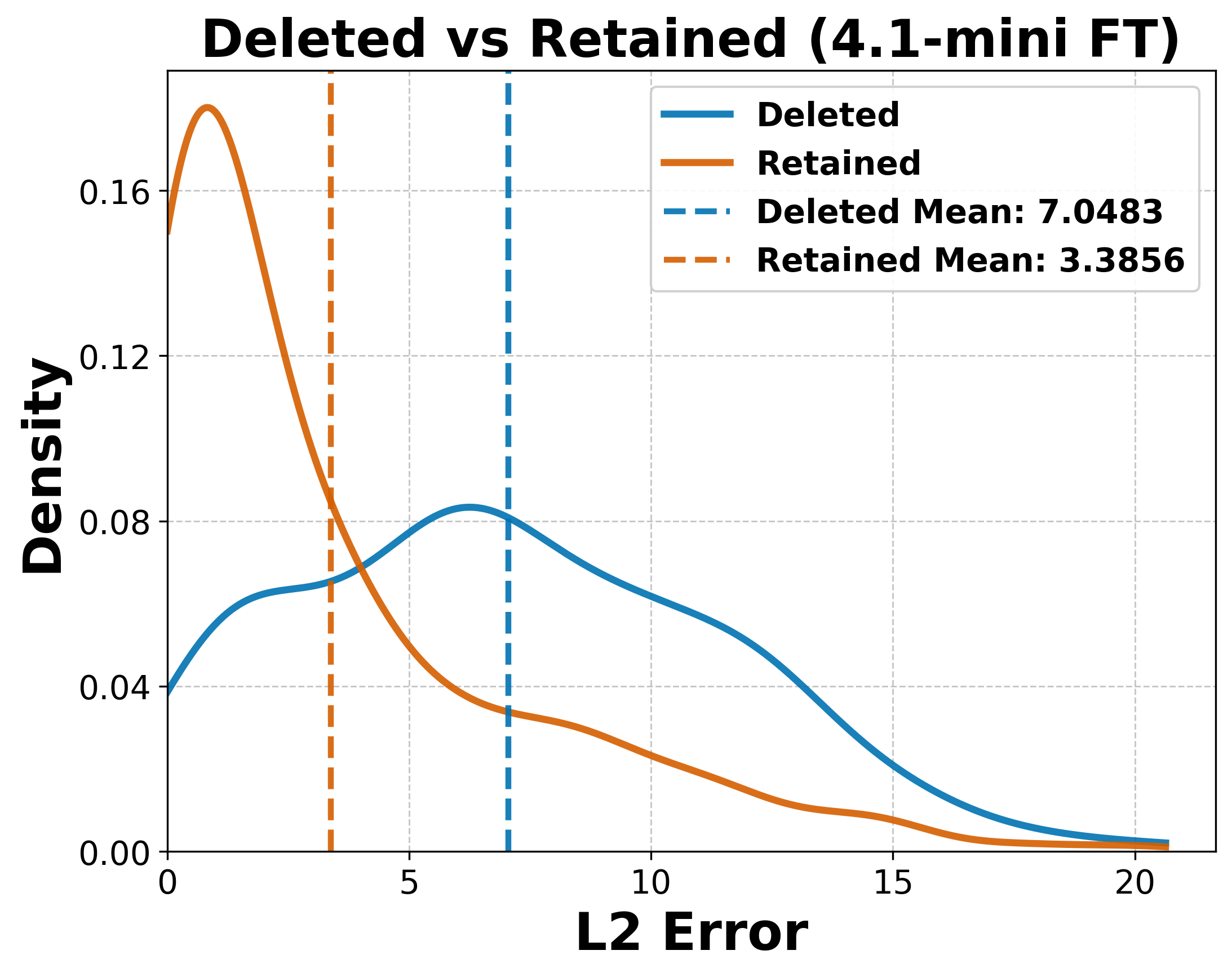}
    \caption{Comparison between deleted records and retained records for AgentDriver. Lower error represents better execution quality.
    }
    \label{fig:appendix_del_retain_agentdriver}
\end{figure}

\begin{table}[h!]
\centering
\caption{Correctness rate (\%) of deleted records and retained records for EhrAgent and CIC-IoT Agent with coarse evaluation.}
\begin{adjustbox}{width=\linewidth} %
\begin{tabular}{c|c|c|c|c}
\toprule
\textbf{Agent} & \textbf{Category} & \textbf{4o-mini} & \textbf{4.1-mini} & \textbf{4.1-mini-FT} \\
\midrule
\multirow{2}{*}{\textbf{EhrAgent}}&Retained & 44.1 & 49.1 & 54.8  \\
&Deleted & 36.3 & 32.1 & 48.2  \\ \midrule
\multirow{2}{*}{\textbf{CIC-IoT}}&Retained & 78.9 & 72.2 & 86.6  \\
&Deleted & 56.7 & 55.1 & 61.0 \\
\bottomrule
\end{tabular}
\end{adjustbox}
\label{tab:appendix_del_retain_ehr}
\end{table}
In this section, we analyze the intrinsic quality differences between retained and deleted memory records.
Figure~\ref{fig:appendix_del_retain_regagent} presents results for the Coarse 1 and Coarse 2 evaluators under history-based deletion, while Figure~\ref{fig:appendix_del_retain_agentdriver} shows results across all evaluators for {AgentDriver}.
Interestingly, when using GPT-4o-mini as the evaluator, the retained memory in {AgentDriver} exhibits lower average quality than the deleted records.
We further report the correctness rates with respect to ground truth for {EhrAgent} and {CIC-IoT Agent}, as shown in Table~\ref{tab:appendix_del_retain_ehr}.

\subsection{Results under memory resource constraints.}
\label{app:differ_size}
We provide the results of the agent running under memory resource constraints in Figure~\ref{fig:agentdriver_ciciot}.
We also provide additional results on memory resource constraints with different memory limitation sizes in Figure~\ref{fig:agentdriver_different_size}.
\begin{figure}[t!]
    \centering
    \includegraphics[width=0.4\textwidth]{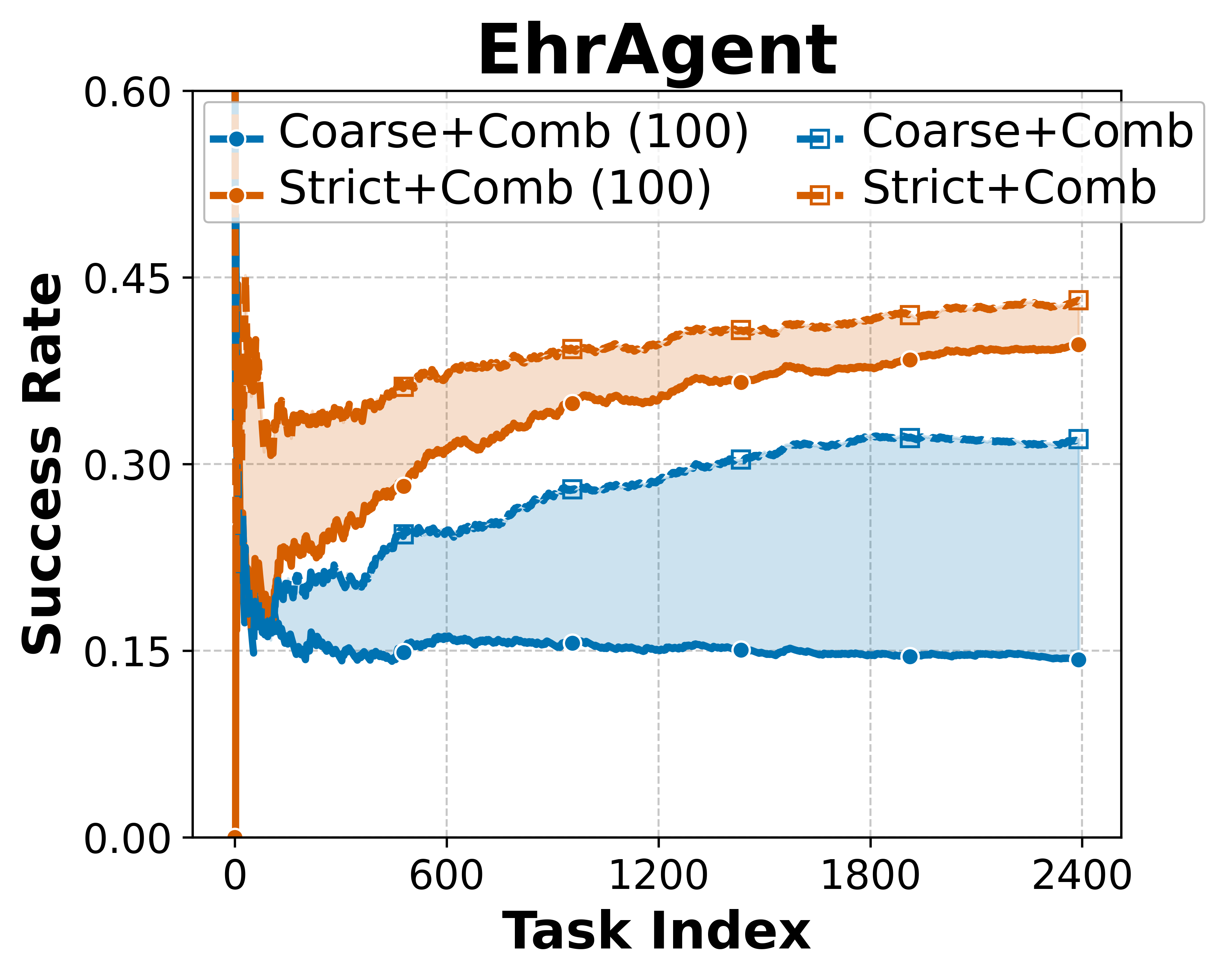}
        \includegraphics[width=0.4\textwidth]{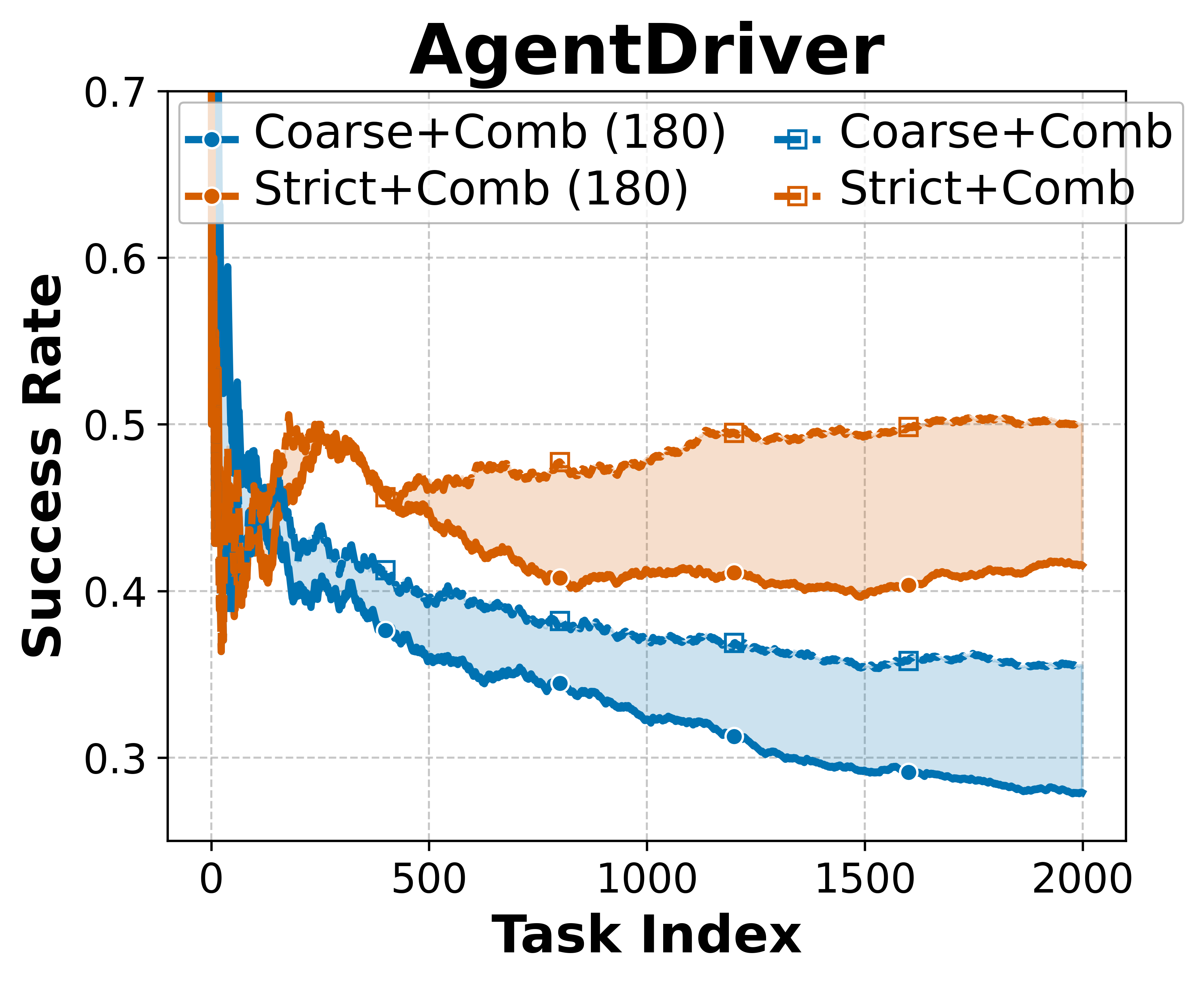}
        \caption{
        Comparison of unlimited memory size versus a limited memory size for strict and coarse selective addition with combined deletion. Coarse here denotes the C1 evaluator.
        }
        \label{fig:agentdriver_ciciot}
    \end{figure}
    \begin{figure}
        \centering
        \centering
  \includegraphics[width=0.4\textwidth]{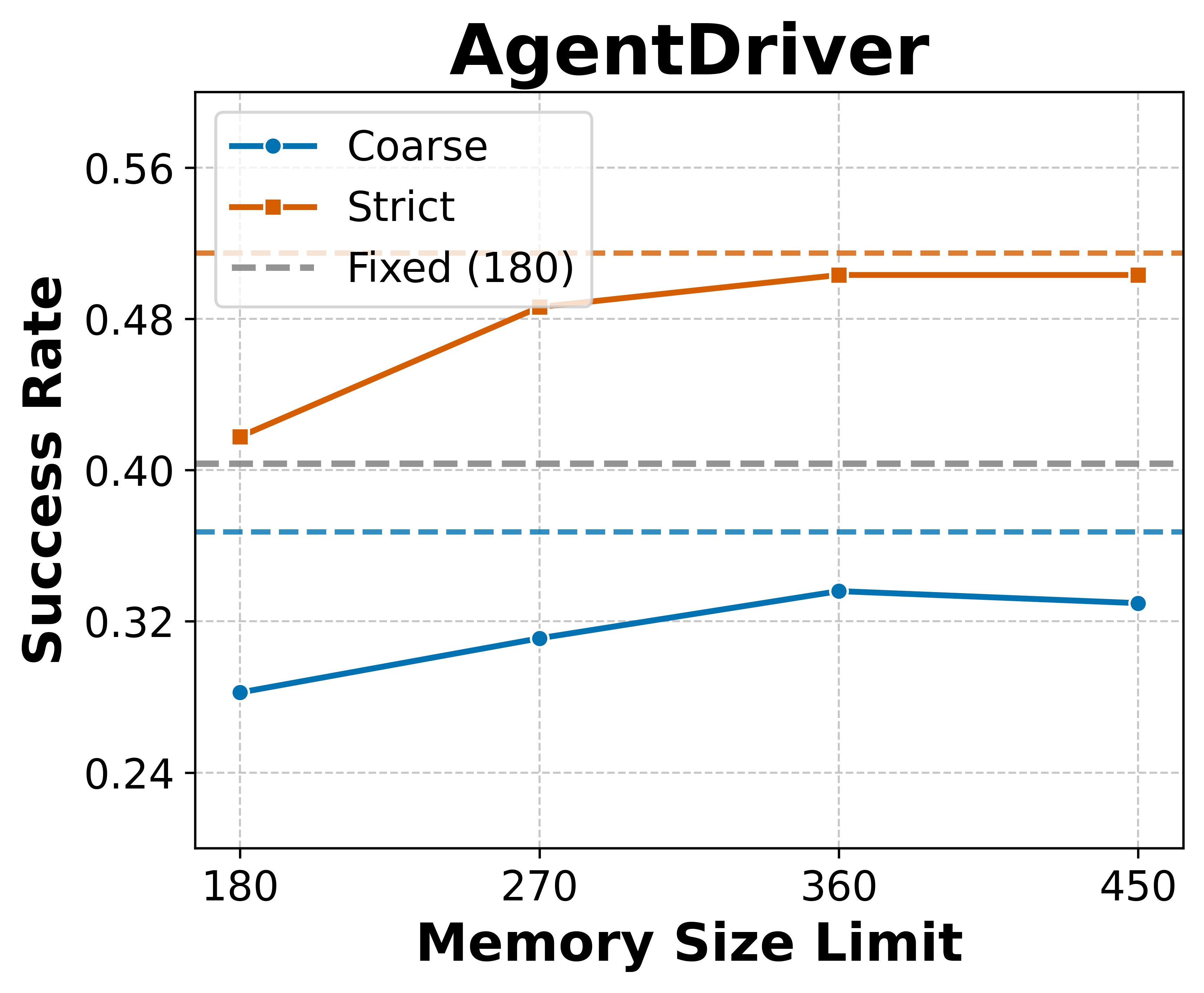}
        \caption{
        Different limited sizes versus the performance of AgentDriver. The horizontal dashed lines are their corresponding unlimited variant performance. Coarse here denotes the C1 evaluator.
        }
        \label{fig:agentdriver_different_size}

    \end{figure}

\end{document}